\documentclass{article} 
\usepackage{sitar_preprint,times}

\usepackage{amsmath,amsfonts,bm}

\def\eqref#1{equation~\ref{#1}}

\def\1{\bm{1}}

\DeclareMathAlphabet{\mathsfit}{\encodingdefault}{\sfdefault}{m}{sl}
\SetMathAlphabet{\mathsfit}{bold}{\encodingdefault}{\sfdefault}{bx}{n}

\usepackage{hyperref}
\usepackage{url}
\usepackage{xspace}

\usepackage{booktabs}
\usepackage{arydshln}
\usepackage{multirow}
\usepackage{graphicx}
\usepackage{xcolor}
\usepackage{tcolorbox}
\newtcolorbox{sitarcase}{colback=black!4,colframe=black!4,
  boxrule=0pt,arc=2pt,left=7pt,right=7pt,top=5pt,bottom=5pt,
  before skip=7pt,after skip=7pt,fontupper=\small}

\usepackage{wrapfig}
\usepackage{needspace}

\newcommand{\name}{\textsc{Sitar}\xspace}

\newcommand{\sitartitlelogo}{\raisebox{-0.28em}[0pt][0pt]{%
  \includegraphics[height=2.0em]{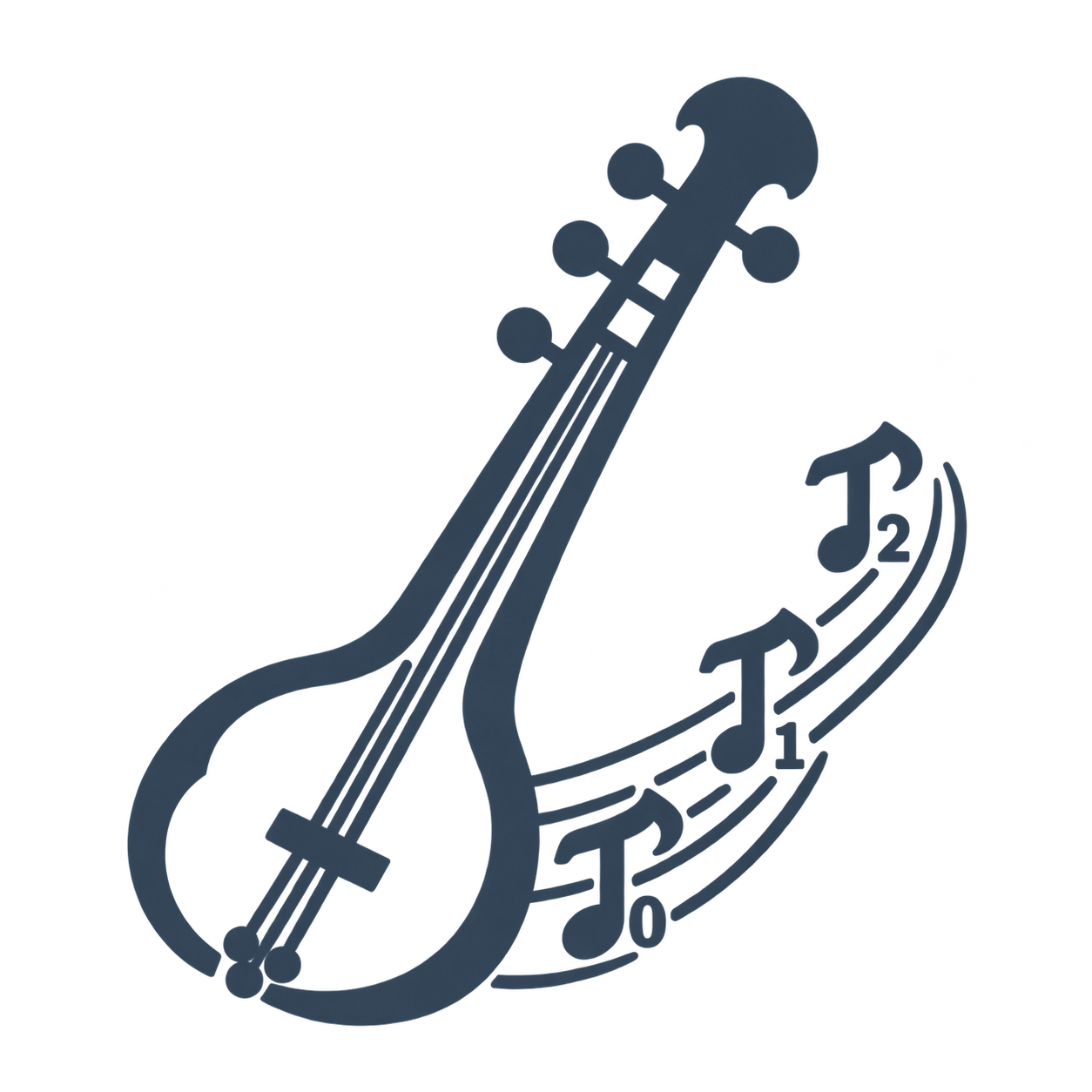}}\hspace{0.3em}}
\title{\sitartitlelogo\name: A Situational Distortion Metric\\
for Automatic Speech Recognition Evaluation}

\newcommand{\sitarauthors}{\parbox{0.96\textwidth}{\centering
\mbox{Mingzhe Lu\textsuperscript{1,2}, Jinhao Liu\textsuperscript{1,2}, Yunpeng Li\textsuperscript{1,2,*}, Hang Su\textsuperscript{3,*}, Lichun Fan\textsuperscript{3}, Jian Luan\textsuperscript{3}}\\[7pt]
{\normalfont\small
\textsuperscript{1}Institute of Information Engineering, Chinese Academy of Sciences, Beijing, China\par
\textsuperscript{2}School of Cyber Security, University of Chinese Academy of Sciences, Beijing, China\par
\textsuperscript{3}Xiaomi Inc.\par
}}}

\author{\sitarauthors}

\iclrfinalcopy
\hypersetup{
  pdftitle={SITAR: A Situational Distortion Metric for Automatic Speech Recognition Evaluation},
  pdfauthor={Mingzhe Lu, Jinhao Liu, Yunpeng Li, Hang Su, Lichun Fan, Jian Luan},
  pdfsubject={Automatic speech recognition evaluation},
  pdfkeywords={speech recognition, evaluation, situational distortion, SITAR},
  hidelinks
}
\begin{document}

\maketitle
\begingroup
\renewcommand{\thefootnote}{\fnsymbol{footnote}}
\footnotetext[1]{\raggedright Corresponding authors:
\href{mailto:liyunpeng@iie.ac.cn}{\texttt{liyunpeng@iie.ac.cn}};\quad
\href{mailto:suhang10@xiaomi.com}{\texttt{suhang10@xiaomi.com}}.}
\endgroup

\begin{abstract}
In automatic speech recognition (ASR), transcription errors with similar lexical or semantic discrepancies can have markedly different consequences for the situation conveyed by speech. Capturing these differences requires not merely contextual assessment, but an understanding of what the spoken discourse describes, establishes, or seeks to bring about. However, relying on LLM judges introduces computational cost and scoring variability. We introduce \name, a situational distortion metric for ASR, realized through a comprehensive end-to-end construction framework. Specifically, this framework aggregates situational severity and cumulative extent into a unified score via a bounded recursive formulation. Using targets derived from this formulation with limited offline LLM support, we train \name as a lightweight BERT-family regressor that scores reference--hypothesis pairs in a single forward pass, entirely free of any prohibitive runtime costs. Recognizing that introducing a new metric demands rigorous justification, we formalize a principled validation protocol spanning both extrinsic validity, assessed at macro and micro levels, and intrinsic scorer reliability. The former, evaluated across diverse datasets and models, demonstrates that \name reliably tracks broad capability scaling while providing orthogonal diagnostic utility, whereas the latter, examined on a purpose-built controlled benchmark, confirms the operational soundness of the scoring instrument itself. Through systematic validation, we demonstrate that \name holds strong potential as a practical, lightweight complement to existing ASR metrics.
We will release the full recipes (code, model weights, dataset, etc.) after the internal review to support community research.
\end{abstract}

\section{Introduction}
Automatic speech recognition (ASR) is pivotal to modern human-computer interaction, yet even small transcription errors can cause downstream failures in conversational voice assistants~\citep{10448230}. Beyond architectural improvements, meaningful progress fundamentally hinges on evaluation metrics that faithfully reflect transcription quality~\citep{aksenova-etal-2021-might}. 
Although modern LLM-based ASR metrics align more closely with human judgments~\citep{parulekar-jyothi-2025-laser}, the prompt-dependent scoring variability~\citep{hong2026rubricsreliablescoresevidencegrounded} and prohibitive computational costs of frequent LLM calls~\citep{huang2026codifyingjudgescalableevaluation} hinder their widespread adoption as standard metrics.
This presents a threefold imperative for next-generation metrics: to effectively harness the understanding capabilities of modern LLMs, to operate independently of them at inference time, and to deliver deterministic scores that make any two transcription errors comparable on a unified scale.
Beyond these operational criteria, and aiming for closer alignment with human judgments, we present a novel target property for transcription evaluation, termed \textbf{situational distortion}.

Across the evolution of ASR metrics, the conceptualization of transcription fidelity has broadened from verbatim correspondence to semantic consistency and, more recently, alignment with human judgments, driving a methodological shift from edit-based error counting toward embedding-based metrics and large language model evaluators. Despite providing a well-defined and deterministic error metric, WER remains indifferent to the meaning carried by language. Subsequent metrics address this limitation at differing levels of granularity, with NE-WER prioritizing micro-level entity components~\citep{garofolo99_accessaudio} and SemDist assessing macro-level semantic representations~\citep{kim21e_interspeech}. However, a gap persists between semantic proximity and alignment with human judgments. For example, a single-token shift from ``depart at fifteen'' to ``depart at fifty'' leaves almost all of the wording unchanged but alters the specified departure time. AER therefore probes comprehension fidelity through task-driven question answering.

Yet, AER~\citep{pulikodan25_interspeech} introduces fresh dilemmas. Although question answering exposes errors invisible to semantic metrics, AER lacks a calibrated scale for comparing transcriptions that lead to failures on entirely different questions and requires repeated LLM queries. LASER~\citep{parulekar-jyothi-2025-laser} also explores fine-tuning a smaller LLM to reduce inference costs. Its rubric assigns fixed penalties to word-level error categories and normalizes their sum by reference length. Although these categories distinguish meaning-preserving from meaning-changing errors, they do not explicitly grade their situational consequences. More fundamentally, although these metrics introduce fine-grained distinctions within their respective basic units, their assessment of the entire reference--hypothesis pair remains a statistical aggregation of locally defined judgments rather than a unified measure of overall transcript distortion.

To develop modern ASR metrics aligned with human judgments, constantly scrambling to address these disparate dilemmas through piecemeal fixes only traps us in treating isolated symptoms. This recognition urges us instead to step back and re-examine the structural foundations of transcription evaluation. Laying these foundations requires treating evaluation as a unified measurement system rather than a collection of heuristics. Specifically, its design must answer four core questions: what property is being measured, against what reference deviations are defined, on what scale their values become comparable, and through what instrument those values are obtained.

Thus, we present \name, an ASR metric that quantifies reference-grounded situational distortion on a common scale and is realized as a lightweight, deterministic scorer. Constructing \name proceeds through these four elements as a coherent sequence, with each establishing the basis for the next. Specifically, we define the \textbf{target property} as the distortion that a hypothesis introduces into the situation conveyed by the spoken source utterance. We establish the \textbf{reference condition} by treating the reference as the gold textual realization of that utterance and taking \(h=r\) as zero distortion, with our proposed Constrained Gap Delineation (CGD) identifying paired gaps as elementary deviation units. These units are first assessed relative to their own reference, yielding reference-specific deviation levels that are then projected onto a \textbf{common scale} according to how they distort the situation conveyed by the spoken source utterance, that is, how they change what that utterance describes, establishes, or seeks to bring about. A bounded recursive formulation subsequently integrates situational severity with cumulative extent into scalar targets. Finally, these targets, constructed offline with LLM-assisted judgments, are used to train the \textbf{operational instrument}, namely \name itself, a lightweight BERT-family regressor requiring no LLM calls at scoring time.

The contributions of this paper can be summarized as follows:
\begin{itemize}
    \item We introduce \name, a lightweight, deterministic ASR metric that quantifies reference-grounded situational distortion in a single forward pass, eliminating any LLM involvement during practical use while achieving closer alignment with human judgments.
    \item We formulate a principled, measurement-grounded methodology to construct \name, establishing an explicit reference anchor, CGD-based deviation delineation, a common scale for reference-specific deviation levels, and a bounded recursive formulation that unifies situational severity with cumulative extent.
    \item We comprehensively validate \name through broad cross-dataset and cross-model evaluations, complemented by a purpose-built controlled benchmark, demonstrating its capacity to track model capability trends, reveal fine-grained diagnostic distinctions, and maintain within-reference severity ordering and cross-reference score consistency.
\end{itemize}

\section{Related Work}
With recent developments in ASR evaluation, we consider it informative to examine literature not only \textbf{methodologically}, regarding how transcription quality is computationally measured and operationalized, but also \textbf{conceptually}, regarding how transcription fidelity is understood and defined.

\textbf{Methodological Evolution.}
At the \emph{lexical} level, Word Error Rate (WER) remains the de facto standard~\citep{morris04_interspeech} despite recognized limitations~\citep{aksenova-etal-2021-might}, with Character Error Rate (CER) as its character-level counterpart~\citep{k-etal-2025-advocating}. Both rely on optimal alignment under uniform edit penalties. However, minimum-edit-distance alignment can be linguistically implausible for splits, merges~\citep{kuhn24_interspeech}, or homophonic spans~\citep{7404808}, motivating relaxed alignment constraints~\citep{11463731}, while uniform penalties overlook error severity, prompting weighted variants based on task relevance~\citep{nanjo05_interspeech} or human perception~\citep{mishra11_interspeech}. Yet these edit-based paradigms remain tied to surface-form discrepancies. Shifting toward semantic preservation, \emph{representational} metrics compare sentence-level embeddings, as in SemDist~\citep{kim21e_interspeech}, while Aligned Semantic Distance~\citep{rugayan22_interspeech} introduces token-level alignment to expose local discrepancies. Representational similarity cannot fully replace lexical signals. BERTScore and WER jointly fit expert ratings of disordered speech better than either alone~\citep{tobin22_s4sg}, supporting hybrid metrics that blend semantic similarity with edit penalties, including $H_{\mathrm{eval}}$~\citep{10389717} and SeMaScore~\citep{sasindran24_interspeech}. Phonetic extensions further target intelligibility~\citep{phukon25_interspeech}. Still, representational similarity alone does not determine practical consequences across diverse contexts, prompting domain-specific weighting for clinically relevant errors~\citep{shor-etal-2023-clinical} and driving the turn toward the \emph{LLM-based} paradigm. Enabled by large-scale pretraining~\citep{DBLP:journals/corr/abs-2001-08361,NEURIPS2022_c1e2faff} and in-context learning~\citep{NEURIPS2020_1457c0d6}, LLM evaluators directly assess error severity, semantic disruption, and acceptable variation, as in LASER~\citep{parulekar-jyothi-2025-laser}. They also support intent-aware probes, with AER~\citep{pulikodan25_interspeech} using question-answering accuracy as a proxy for fidelity and $\text{S}^2\text{ER}$~\citep{jiang2026humanlikeinteractivespeechrecognition} evaluating the preservation of core intent and meaning-bearing content. Nevertheless, prompt-based scoring introduces instability across rubric perturbations~\citep{hong2026rubricsreliablescoresevidencegrounded} and repeated runs~\citep{haldar-hockenmaier-2025-rating}, while prohibitive inference latency and computational cost restrict scalability, motivating selective invocation of LLM judges~\citep{huang2026codifyingjudgescalableevaluation}.

\textbf{Conceptual Perspectives.}
Under the \emph{surface} view, fidelity is defined through strict verbatim correspondence, yet lexical discrepancies do not consistently translate into downstream comprehension failures. Lower slot-understanding error can coexist with higher WER~\citep{1318504}, and errors involving key actions or entities can prevent a voice assistant from fulfilling a request even when other lexical errors leave the intended action intact~\citep{10448230}. In practical decision audits, WER poorly tracks how effectively humans extract decisions from transcripts~\citep{favre13_interspeech}. The \emph{semantic} view therefore assesses whether asserted or implied content is preserved, drawing on semantic textual similarity~\citep{agirre-etal-2012-semeval} and natural language inference~\citep{bowman-etal-2015-large}. However, equivalent propositional content can function differently across discourse contexts or conversational positions. In clinical dialogue, standard surface and semantic metrics correlate poorly with clinician-assigned risk labels~\citep{ellis-etal-2026-wer}. This motivates our \emph{situational} perspective, grounded in cognitive situation models where comprehenders integrate events and entities into an evolving representation of the described state of affairs~\citep{10.1037/0033-2909.123.2.162}, as well as communicative grounding through which interlocutors establish mutual understanding sufficient for their current purposes~\citep{10.1037/10096-006}. Bridging these cognitive and interactive views, we define the situation relevant to ASR evaluation in terms of what the spoken source utterance describes, establishes, or seeks to bring about, and evaluate transcription errors by the degree to which they distort this situation.

\section{Methodology}
\label{sec:methodology}
\begin{figure}[t]
    \centering
    \includegraphics[width=0.98\linewidth]{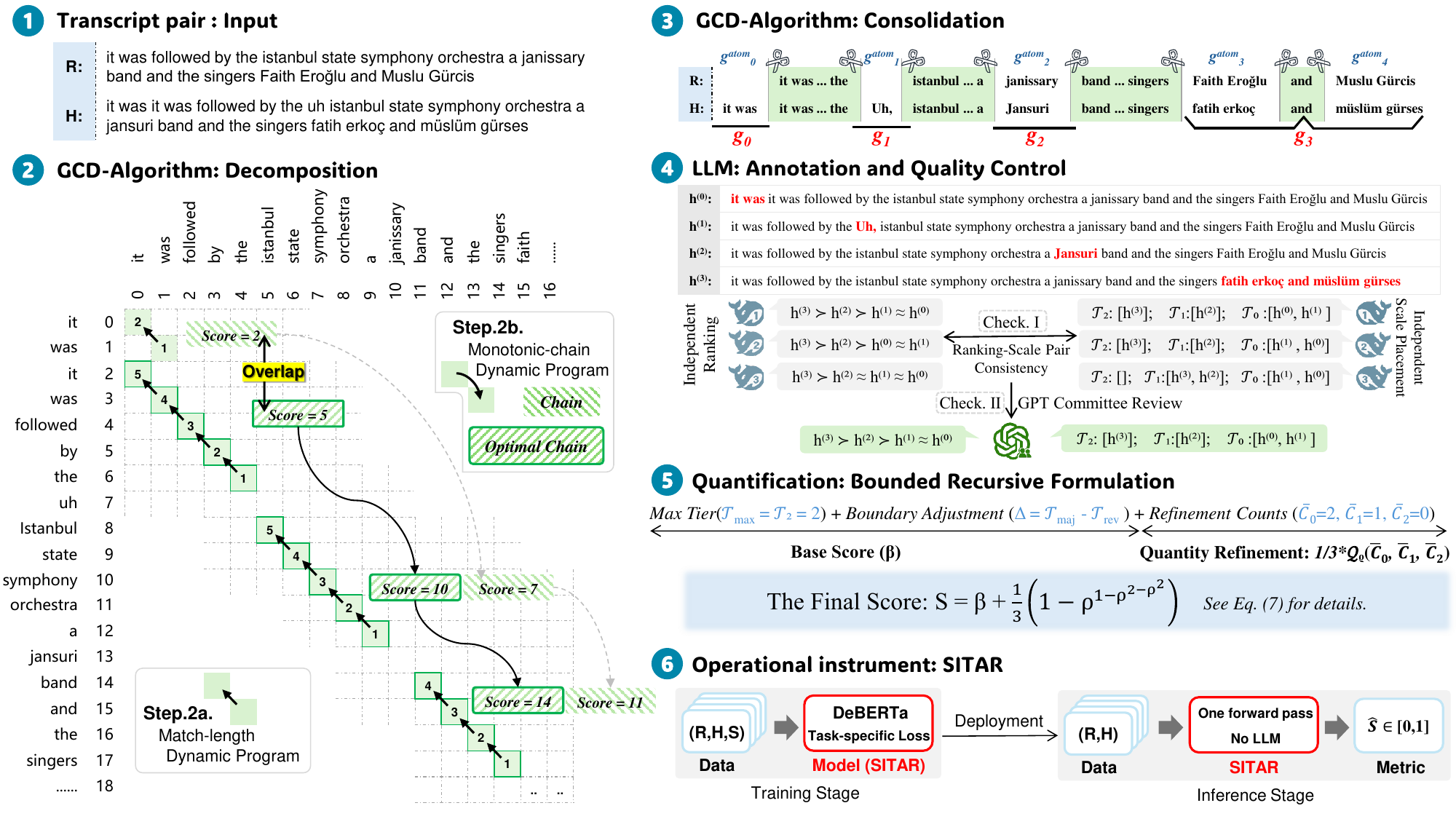}
    \caption{Overview of \name construction: CGD extracts paired gaps from transcripts, which are annotated for situational severity and aggregated into supervision targets to train the LLM-free scorer.}
    \label{fig:sitar-framework}
\end{figure}

Let $r \in \mathcal{R}$ denote the reference transcript of the raw audio and $h \in \mathcal{H}$ the ASR hypothesis. Figure~\ref{fig:sitar-framework} exemplifies the operational workflow of \name on a concrete instance, with each procedural stage detailed below.
\subsection{Reference Condition}
The primary prerequisite for reliable scoring is establishing an unambiguous, well-bounded unit of comparison across a reference $r$ and a hypothesis $h$. Its boundaries must respect the informational structure of the source utterance, rendering word-level Levenshtein alignment ill-suited for defining such units~\citep{7404808,11463731}. While LLMs can delineate such units, costly inference limits scalability and variable outputs undermine reproducibility. 
We therefore introduce \textbf{Constrained Gap Delineation (CGD)}, a deterministic algorithm that locates text discrepancies using shared text as anchors and consolidates adjacent differences into paired gaps $g_i$. Each gap pairs an offset-defined character span from $r$ with one from $h$, either of which may be empty, and serves as an elementary unit for subsequent situational scoring.
We begin by normalizing \(r\) and \(h\) into word sequences \(\tilde{r}\) and \(\tilde{h}\), respectively, through lowercasing and punctuation removal, while retaining mappings from normalized tokens to their original character spans. In what follows, we elaborate on the algorithmic details of CGD.

\textbf{Decomposition.}
To identify anchors, we recursively process paired regions $(u,v)$, beginning with $(\tilde{r},\tilde{h})$. Within each region, we set $n_u=|u|$ and $n_v=|v|$ and identify maximal common word blocks extension using a \emph{match-length dynamic program}: $L_{p,q}$ denotes the number of consecutive matches starting at $u_p$ and $v_q$, with $L_{p,q}=1+L_{p+1,q+1}$ if $u_p=v_q$ and $0$ otherwise, for $0\leq p<n_u$ and $0\leq q<n_v$. The boundary values are $L_{n_u,q}=0$ for $0\leq q\leq n_v$ and $L_{p,n_v}=0$ for $0\leq p\leq n_u$. Requiring left-maximality with a minimum block length $\ell_{\min}$ gives the candidate set
\begin{equation}
\label{eq:cgd-anchor-candidates}
\mathcal B
=
\left\{
(p,q,L_{p,q}) :
L_{p,q}\geq\ell_{\min},\;
p=0\ \lor\ q=0\ \lor\ u_{p-1}\neq v_{q-1}
\right\}.
\end{equation}
We set $\ell_{\min}=2$ to suppress spurious isolated matches, especially on function words such as \emph{the} and \emph{to}, and recompute $\mathcal B$ with $\ell_{\min}=1$ if it is empty. If $\mathcal B$ remains empty, no anchor is selected for the current region. Otherwise, we sort the candidates by $(p,q,-\ell)$ into indexed triples $(p_i,q_i,\ell_i)$. A second \emph{monotonic-chain dynamic program} then selects an ordered, nonoverlapping chain with the greatest cumulative matched length. Let $f_i$ denote the maximum cumulative matched length of such a chain ending at candidate $i$:
\begin{equation}
\label{eq:cgd-anchor-chain}
f_i = \ell_i + \max\left(
\left\{f_j : j<i,\;
p_j+\ell_j\leq p_i,\;
q_j+\ell_j\leq q_i
\right\}
\cup \{ 0 \}
\right).
\end{equation}
The optimal chain is recovered by backtracking from the earliest index in $\arg\max_i f_i$, with predecessor ties likewise resolved in favor of the earliest candidate in the sort order.

We then filter the matching word-position pairs $(p,q)$ in the selected chain using forward and backward edit-distance dynamic programs. Let $E^+_{a,b}$ and $E^-_{a,b}$ denote the resulting unit-cost prefix and suffix edit distances for $(u[:a],v[:b])$ and $(u[a:],v[b:])$, respectively. Both use unit costs for insertion and deletion and substitution cost $\delta_{a,b}=\mathbf{1}[u_a\neq v_b]$, with boundary values $E^+_{a,0}=a$, $E^+_{0,b}=b$, $E^-_{a,n_v}=n_u-a$, and $E^-_{n_u,b}=n_v-b$. Because a matching pair satisfies $u_p=v_q$, it lies on a minimum-cost edit path if and only if
\begin{equation}
\label{eq:cgd-anchor-filter}
E^+_{p,q}
+
E^-_{p+1,q+1}
=
E^+_{n_u,n_v}.
\end{equation}
Retained matching pairs contiguous in both sequences form anchor blocks. We recursively repeat this selection and filtering in the paired subregions before, between, and after these blocks. Recursion terminates in subregions with no retained matches; terminal subregions with at least one nonempty side form the ordered atomic gaps
$g_t^{\mathrm{atom}}=(\tilde{r}[I_t^r],\tilde{h}[I_t^h])$,
where $I_t^r$ and $I_t^h$ are zero-based word intervals in $\tilde{r}$ and $\tilde{h}$, respectively.

\textbf{Consolidation.}
Atomic gaps can over-segment a coherent local deviation when opportunistic matches split it into multiple fragments. CGD therefore complements lexical matching with linguistic constraints, consolidating adjacent gaps into elementary deviation units. Specifically, given atomic gaps $g_0^{\mathrm{atom}},\ldots,g_{M-1}^{\mathrm{atom}}$, we identify those with at most three tokens per side as candidate fragments, and examine whether adjacent gaps can merge with them into an elementary deviation unit. Such consolidation is permitted only when intervening matches do not cross informational boundaries and the resulting envelope spans at most one predicate per side. For example, matching the shared word \emph{search} splits a coherent local deviation into two atomic gaps, which CGD consolidates:
\[
\begin{aligned}
r:\;& \text{Okay, so I'm just }\underbrace{\overbrace{\text{gonna}}^{g_i^{\mathrm{atom}}}\ \text{search}\ \overbrace{\text{the app}}^{g_{i+1}^{\mathrm{atom}}}}_{\text{elementary deviation unit}}\text{ so I can see that we do have a flight.}\\
h:\;& \text{Okay, so I'm just }\underbrace{\overbrace{\text{going to}}^{g_i^{\mathrm{atom}}}\ \text{search}\ \overbrace{\text{that up}}^{g_{i+1}^{\mathrm{atom}}}}_{\text{elementary deviation unit}}\text{ so I can see that we do have a flight.}
\end{aligned}
\]
Let $\mathcal A_i$ denote the endpoint indices $j\in\{i,\ldots,M-1\}$ for which $g_i^{\mathrm{atom}},\ldots,g_j^{\mathrm{atom}}$ forms an admissible group, where singleton feasibility guarantees $i\in\mathcal A_i$. CGD then computes a minimum-cardinality partition of the atomic-gap sequence by dynamic programming over suffixes. Let $d_i$ denote the minimum number of groups covering $g_i^{\mathrm{atom}},\ldots,g_{M-1}^{\mathrm{atom}}$. With $d_M=0$, the recurrence is
\begin{equation}
\label{eq:cgd-partition}
d_i := 1+\min_{j\in\mathcal A_i}d_{j+1}, \qquad i=M-1,\ldots,0.
\end{equation}
The optimal partition is decoded forward from $i=0$, with ties broken in favor of the largest endpoint. Retained character-offset mappings then project each consolidated group back to the original transcripts, yielding the ordered paired gaps $g_0,\ldots,g_{m-1}$.

For an original transcript pair \((r,h)\), let \(g_0,\ldots,g_{m-1}\) denote the paired gaps identified by CGD. For each \(g_i\), we construct a complete single-gap hypothesis \(h^{(i)}\). It retains the hypothesis-side content of \(g_i\), while all other gaps are restored with their reference-side content. This produces \(m\) single-gap pairs, \((r,h^{(0)}),\ldots,(r,h^{(m-1)})\), all sharing the same reference.

\subsection{Common Scale}
The distortion caused by each $g_i$ reflects how strongly $h^{(i)}$ alters the information conveyed by $r$. When $h^{(i)}$ induces less distortion than $h^{(j)}$, we denote $g_i \prec^{(r)} g_j$. When their impacts are approximately equal, we denote $g_i \approx^{(r)} g_j$. Anchoring this relation at the zero-deviation baseline \(g_\emptyset\) yields a reference-specific ordering of the schematic form:
\begin{equation}
\label{eq:contextual-gap-order}
g_\emptyset \prec^{(r)} \cdots \prec^{(r)} g_i \prec^{(r)} \cdots \prec^{(r)} g_j \approx^{(r)} g_k \prec^{(r)} \cdots,
\qquad \text{where } g_\emptyset \leftrightarrow \bigl(h^{(\emptyset)}\equiv r\bigr).
\end{equation}

However, a general metric must make transcription deviations comparable across hypotheses produced by any ASR model for any reference. Although existing metrics, exemplified by lexical edit distance and representational similarity, yield common numerical scales, their implicit priors impose an intrinsic ceiling on evaluation, restricting each metric to a partial view of the situation conveyed by the reference. Any common scale necessarily abstracts, but where that abstraction stops determines the highest level of change the scale can register, as in WER, whose defining advantage in lexical fidelity serves as the very barrier blinding it to higher-level semantic consequences.

We seek comparability not by impoverishing the object of measurement, but through a shared structure capable of preserving effects across distinct levels. Results-chain frameworks\footnote{OECD, \href{https://www.oecd.org/en/publications/development-co-operation-tips-tools-insights-practices_be69e0cf-en/managing-for-sustainable-development-results_8e326a5d-en.html}{\emph{Managing for Sustainable Development Results}}, 2024.} provide such a structure by organizing results into linked tiers and treating the links among them as integral to evaluation. Specifically, they distinguish three progressively broader levels, namely changes confined to the immediate result, changes that alter subsequent outcomes, and changes that lead to higher-level consequences, whether intended or unintended. When it comes to a spoken utterance, its effects may arise through what it describes, establishes, or seeks to bring about. 

Across these functions, we argue that transcription deviations can be distinguished by whether they remain confined to expression, alter uptake without changing the intended outcome, or affect the ultimate outcome. These distinctions lead us to define three foundational consequence tiers, formalizing an evaluation perspective we term \textbf{situational distortion}. At $\mathcal{T}_0$, the situation is preserved despite differences in expression. At $\mathcal{T}_1$, uptake changes while the intended outcome is preserved. At $\mathcal{T}_2$, the ultimate outcome changes, including consequences for decisions and execution. These tiers are deliberately coarse-grained to reduce ambiguity in severity assignment. Finer resolution is subsequently introduced through directional boundary cases and tier-wise gap counts, yielding a composite score that captures both the severity and extent of distortion.

We assign nominal scores \(0,1,2\) to \(\mathcal T_0,\mathcal T_1,\mathcal T_2\), respectively. Between adjacent tiers \(\mathcal T_i\) and \(\mathcal T_{i+1}\), we distinguish two directional boundary cases: \(\mathcal T_i\rightarrow\mathcal T_{i+1}\) and \(\mathcal T_{i+1}\rightarrow\mathcal T_i\). Here, the starting tier gives the gap's nominal assignment, while trisecting the unit interval gives a \(1/3\) severity adjustment in the arrow's direction. For a transcript pair with at least one gap, the highest severity position among its gaps defines the base score \(\beta\):
\begin{equation}
    \label{eq:tier-derived-base}
    \beta\in\left\{0,\,0+1/3,\,1-1/3,\,1,\,1+1/3,\,2-1/3,\,2\right\}.
\end{equation}
Each \(\beta\) anchors a nonoverlapping interval \([\beta,\beta+1/3)\) reserved for refinement by cumulative extent, yielding an overall scale of \([0,2+1/3)\).

To capture cumulative extent, let \(c_k\) count gaps assigned to nominal tier \(\mathcal T_k\), for \(k\in\{0,1,2\}\). Here, \(k_{\max}=\max\{k:c_k>0\}\) identifies the nominal tier of the most severe gap, whereas \(\beta\) records its exact severity position, including any directional boundary adjustment. Quantity refinement counts gaps by nominal tier irrespective of boundary adjustments, to avoid unnecessary combinatorial complexity. Since \(\beta\) already accounts for one gap in \(\mathcal T_{k_{\max}}\), refinement begins with the second gap at that tier and includes all gaps at lower tiers, yielding counts \(\bar c_k=c_k-\mathbf{1}[k=k_{\max}]\). Within each base-score interval, this refinement preserves tier priority: of two distinct count configurations, the one with more gaps at the highest tier where they differ receives the higher score. For a fixed parameter $0<\rho<1$ controlling saturation, we define the score as
\begin{equation}
    \label{eq:bounded-scalarization}
    S(r,h)=\beta+(1/3)\mathcal Q_\rho\bigl(\bar c_0,\ldots,\bar c_{k_{\max}}\bigr).
\end{equation}
Here, $\mathcal Q_\rho$ aggregates counts recursively from lower to higher tiers, with $x_j$ denoting the nonnegative integer refinement count at tier $j$:
\begin{equation}
    \label{eq:recursive-tier-aggregation}
    \mathcal Q_\rho(\emptyset)=0,\qquad \mathcal Q_\rho(x_0,\ldots,x_k)=1-\rho^{\,x_k+\mathcal Q_\rho(x_0,\ldots,x_{k-1})}.
\end{equation}
In particular, \(\mathcal Q_\rho(x_0)=1-\rho^{x_0}\). For finite counts, \(0\leq\mathcal Q_\rho<1\), ensuring that \(S(r,h)\in[\beta,\beta+1/3)\). This bound, combined with integer counts and monotone recursive updates, preserves tier priority while allowing lower-tier gaps to refine the score. A notable case is benign speech disfluencies, where multiple $\mathcal{T}_0$ gaps yield a positive but saturating refinement, yet preserve the situation. Appendix~\ref{app:score-construction} provides the parameter derivation, proofs, and worked examples.

\subsection{Operational Instrument: \name}
In practice, we first rank the single-gap pairs $(r,h^{(i)})$ to establish a reference-specific ordering of the form shown in Eq.~\ref{eq:contextual-gap-order}, and subsequently assign each a severity tier independently. Decoupling these steps provides a self-consistency check within each round, and any contradiction between the ranking and tier assignments in any round leads to rejection of the entire transcript pair. Specifically, both the ranking and tier annotations are generated by DeepSeek-V4-Pro across three independent rounds to mitigate stochasticity.

At this stage, each single-gap hypothesis $h^{(i)}$ receives three candidate tier assignments generated by DeepSeek, $\{t_{\text{DS}}^{(i,k)}\}_{k=1}^3$. For each $h^{(i)}$, we obtain the majority tier $t_{\text{maj}}^{(i)}$ across the three rounds; if any hypothesis in a transcript pair lacks such a majority, the entire pair is discarded. Qualified pairs then undergo quality review by a four-agent committee using GPT-5.6-Sol-Max in ultra mode, which assigns a reviewed tier $t_{\text{rev}}^{(i)}$ to each hypothesis. We apply a second quality filter, discarding the pair if any constituent hypothesis violates the consistency condition
\begin{equation}
    \label{eq:annotation-consistency}
    \lvert t_{\text{maj}}^{(i)} - t_{\text{rev}}^{(i)} \rvert \le 1 \quad \land \quad \left( t_{\text{dis}}^{(i)} \text{ is undefined} \;\lor\; \lvert t_{\text{maj}}^{(i)} - t_{\text{dis}}^{(i)} \rvert \le 1 \right),
\end{equation}
where $t_{\text{dis}}^{(i)}$ denotes the unique dissenting DeepSeek assignment if one exists.

For each retained hypothesis, $t_{\text{rev}}^{(i)}$ sets the nominal score, while the sign of $t_{\text{maj}}^{(i)} - t_{\text{rev}}^{(i)}$ determines the directional boundary adjustment. The maximum adjusted position within each transcript pair defines its base score $\beta$. After excluding one hypothesis that attains this maximum, we count the remaining hypotheses by their GPT-reviewed tiers to obtain the refinement counts $\bar c_k$. Substituting $\beta$ and these refinement counts into Eq.~\ref{eq:bounded-scalarization} yields the deterministic supervision target $S(r,h)$.

For scalable, LLM-free evaluation with low latency, we amortize this multi-stage pipeline into a lightweight regressor $f_\theta(r,h) \in \mathbb{R}$. Specifically, we implement $f_\theta$ using DeBERTa-v3-large~\citep{DBLP:conf/iclr/HeGC23} and optimize it on reference-hypothesis pairs using exclusively the scalar supervision targets $S(r,h)$ via three complementary objectives:
\begin{equation}
    \mathcal{L}(\theta)
    =
    \lambda_{\mathrm{mse}}\mathcal{L}_{\mathrm{mse}}
    + \lambda_{\mathrm{boundary}}\mathcal{L}_{\mathrm{boundary}}
    + \lambda_{\mathrm{ordinal}}\mathcal{L}_{\mathrm{ordinal}}.
\end{equation}
In this joint objective, $\mathcal{L}_{\mathrm{mse}}$ regresses toward the target scores, while $\mathcal{L}_{\mathrm{boundary}}$ penalizes predictions outside the score interval containing the target. Additionally, $\mathcal{L}_{\mathrm{ordinal}}$~\citep{Niu_2016_CVPR} supervises an auxiliary head to predict whether the target score reaches each predefined threshold. During inference, $f_\theta$ evaluates any reference-hypothesis pair in a single forward pass, and we clip and normalize the raw prediction to $[0,1]$ such that lower values indicate less situational distortion:
\begin{equation}
\label{eq:normalized-scorer}
\widehat{S}(r,h)
=
\frac{\operatorname{clip}\left(f_\theta(r,h),\,0,\,7/3\right)}{7/3}.
\end{equation}


\section{Experiments}
To justify a new evaluation metric alongside established baselines, we evaluate \name along two complementary axes of \textbf{extrinsic validity}. At the macro level, the metric must faithfully track broad developmental progress with systematically decreasing distortion. At the micro level, it must offer orthogonal diagnostic insights, differentiating nuanced error behaviors that conventional metrics treat identically. Finally, we evaluate the \textbf{intrinsic reliability} of the trained \name regressor itself.

\subsection{Benchmark Setup}
\textbf{\name{}-specific.} The supervision corpus originates primarily from AppTek~\cite{beck26_interspeech}, with transcripts generated by four ASR models. Through quality filtering, FLEURS error augmentation, and selective gap-correction synthesis on multi-gap instances, we construct a supervision dataset of 12,347 pairs. Data construction and scorer training are detailed in Appendices~\ref{app:training-data-construction} and~\ref{app:scorer-construction}, respectively.

\textbf{Cross-domain.} We evaluate \name across domains and model families using the FLEURS English test split~\citep{10023141}, LibriSpeech~\citep{7178964}, and VoxPopuli English~\citep{wang-etal-2021-voxpopuli}, with five Whisper variants~\citep{pmlr-v202-radford23a} ranging from tiny to large alongside the base and large variants of wav2vec~2.0~\citep{NEURIPS2020_92d1e1eb}. Among these datasets and models, only a portion of the FLEURS training split and selected erroneous Whisper-large-v3 transcriptions contributed to supervision; LibriSpeech, VoxPopuli, and outputs from the other evaluated models were excluded. We compare against five baselines: WER and Named Entity WER, abbreviated as NE-WER~\citep{garofolo99_accessaudio}, for lexical overlap; BERTScore$_{\mathrm{F1}}$~\citep{DBLP:conf/iclr/ZhangKWWA20} and SemDist~\citep{kim21e_interspeech} for semantic similarity; and AER for task-driven comprehension. For AER, we use the scores reported by \citet{pulikodan25_interspeech}.

\begin{table}[t]
\centering
\caption{Comparison of ASR evaluation metrics across datasets.}
\label{tab:system-level-comparison}
\resizebox{\textwidth}{!}{%
\begin{tabular}{cccccccc}
\toprule
Model Name & Dataset & WER & NE-WER & BERTScore$_{\mathrm{F1}}$ & SemDist & AER & \name \\
\midrule
whisper-tiny   & \multirow{7}{*}{fleurs\_en}
              & 0.1446 & 0.6500 & 0.9437 & 0.0711 & 0.2478 & 0.3872 \\
whisper-base  & & 0.1150 & 0.5500 & 0.9516 & 0.0558 & 0.2015 & 0.2999 \\
whisper-small & & 0.0711 & 0.4250 & 0.9585 & 0.0457 & 0.1586 & 0.2113 \\
whisper-medium & & 0.0671 & 0.3167 & 0.9610 & 0.0421 & 0.1508 & 0.1590 \\
whisper-large & & 0.0465 & 0.3000 & 0.9621 & 0.0410 & 0.1477 & 0.1388 \\
\cmidrule{1-1}\cmidrule{3-8}
wav2vec2-base  & & 0.2016 & 0.7750 & 0.8709 & 0.0855 & 0.2393 & 0.3670 \\
wav2vec2-large & & 0.1719 & 0.7917 & 0.8752 & 0.0651 & 0.2301 & 0.3247 \\
\midrule
whisper-tiny   & \multirow{7}{*}{librispeech}
              & 0.0816 & 0.2348 & 0.8890 & 0.0828 & 0.3009 & 0.2257 \\
whisper-base  & & 0.0558 & 0.1927 & 0.8912 & 0.0776 & 0.2897 & 0.1631 \\
whisper-small & & 0.0397 & 0.1569 & 0.8927 & 0.0738 & 0.2600 & 0.1047 \\
whisper-medium & & 0.0324 & 0.1356 & 0.8927 & 0.0729 & 0.2637 & 0.0818 \\
whisper-large & & 0.0233 & 0.1199 & 0.8924 & 0.0672 & 0.2540 & 0.0634 \\
\cmidrule{1-1}\cmidrule{3-8}
wav2vec2-base  & & 0.0356 & 0.1205 & 0.9902 & 0.0183 & 0.2696 & 0.1585 \\
wav2vec2-large & & 0.0290 & 0.1136 & 0.9918 & 0.0149 & 0.2643 & 0.1344 \\
\midrule
whisper-tiny   & \multirow{7}{*}{voxpopuli}
              & 0.1387 & 0.3580 & 0.9583 & 0.0482 & 0.4723 & 0.3157 \\
whisper-base  & & 0.1149 & 0.2716 & 0.9628 & 0.0397 & 0.4505 & 0.2532 \\
whisper-small & & 0.1025 & 0.1481 & 0.9660 & 0.0346 & 0.4567 & 0.1984 \\
whisper-medium & & 0.0920 & 0.1605 & 0.9675 & 0.0312 & 0.4739 & 0.1701 \\
whisper-large & & 0.1084 & 0.1605 & 0.9651 & 0.0406 & 0.4345 & 0.1684 \\
\cmidrule{1-1}\cmidrule{3-8}
wav2vec2-base  & & 0.2247 & 0.7407 & 0.8754 & 0.1334 & 0.4387 & 0.4084 \\
wav2vec2-large & & 0.1876 & 0.6914 & 0.8799 & 0.1017 & 0.4366 & 0.3577 \\
\bottomrule
\end{tabular}%
}
\end{table}

\subsection{Extrinsic Metric Validity}
\textbf{Macro-Level.}
Intuitively, as an ASR system generates more accurate transcriptions, evaluation metrics should broadly reflect this improvement. Model scaling serves as a natural testbed, leading us to evaluate Whisper across varying sizes. As capacity expands from tiny to large, transcription performance tends to improve, as reflected by existing metrics such as WER, SemDist, and AER in Table~\ref{tab:system-level-comparison}. \name follows the same trend across all three benchmarks and both Whisper and wav2vec2. These evaluations include LibriSpeech and VoxPopuli, which were absent from its supervision data, alongside Whisper variants from tiny to medium and wav2vec2, whose outputs were not used for supervision. For example, on VoxPopuli, \name decreases from 0.4084 for wav2vec2-base to 0.3577 for wav2vec2-large.

\textbf{Micro-Level.}
Beyond macro-level trends, a new metric should provide diagnostic utility orthogonal to established baselines. Conventional metrics often occupy narrow ranges for modern ASR systems, with WER and SemDist below 0.1 and BERTScore${\mathrm{F1}}$ around 0.9--0.99. \name scores span 0.06 to 0.40, showing greater dispersion similar to NE-WER and AER. Table~\ref{tab:system-level-comparison} further reveals rank inversions across model families. On LibriSpeech, \name favors whisper-small (0.1047) over wav2vec2-large (0.1344), reversing the ordering under WER, NE-WER, BERTScore${\mathrm{F1}}$, and SemDist. On FLEURS, whisper-tiny leads wav2vec2-large under WER, NE-WER, and BERTScore$_{\mathrm{F1}}$, but receives a higher \name score (0.3872 versus 0.3247). Comprehension-centric AER agrees with \name in both cases, as does SemDist on FLEURS. However, AER's uncached evaluation requires four test-time LLM calls per transcript pair~\citep{pulikodan25_interspeech}, whereas \name requires none. Appendix~\ref{app:aer-comparison} provides further comparative analysis.

\subsection{Intrinsic Reliability}
\label{sec:intrinsic-reliability}
We next evaluate the intrinsic quality of \name using 330 controlled transcript pairs in $\mathcal{D}_{\mathrm{ctrl}}$, comprising 11 hypothesis variants for each of 30 English references across finance, culture and sports, public affairs, services, and general communication. The construction profiles are listed in Table~\ref{tab:intrinsic-score-structure}. Appendix~\ref{app:human-annotation} provides a complete construction example, annotation instructions, and analysis details, including the prediction-based selection of the 30 references from 35 candidates.

\textbf{Validating the Common Scale.}
A common scale should map comparable error structures to consistent severity regions across contexts. Grouping by the planned tier and gap-count profiles, Table~\ref{tab:intrinsic-score-structure} summarizes normalized score distributions across references in $\mathcal{D}_{\mathrm{ctrl}}$. These groups retain their construction assignments; the subsequent ranking analysis uses final human-adjudicated labels, including the correction detailed in Appendix~\ref{app:human-agreement-details}. Expression-only profiles $\mathcal{T}_0$ and $2\mathcal{T}_0$ receive consistently low scores across references, with medians below 0.10 and interquartile ranges near 0.04. For the $\mathcal{T}_1$ and $\mathcal{T}_2$ profiles, distribution centers shift upward, with greater dispersion and partially overlapping central intervals. Profiles containing both tiers have medians above 0.70 and narrower interquartile ranges, indicating higher score positions with less dispersion. Within a fixed planned $\mathcal{T}_1$ and $\mathcal{T}_2$ composition, adding $\mathcal{T}_0$ gaps produces small changes in location and spread. For example, adding two $\mathcal{T}_0$ gaps to $\mathcal{T}_2+\mathcal{T}_1$ increases the mean by 0.0030 and standard deviation by 0.0082, consistent with expression-level differences providing modest refinements to principal severity position.

\begin{table}[h]
\centering
\caption{Cross-reference distributions of normalized \name scores.}
\label{tab:intrinsic-score-structure}
\begin{tabular}{lrrrrr}
\toprule
Construction profile & Mean & Median & Standard deviation & Central 50\% & IQR \\
\midrule
$\mathcal{T}_0$
& 0.0813 & 0.0763 & 0.0491 & [0.0575, 0.0986] & 0.0411 \\
$2\mathcal{T}_0$
& 0.1268 & 0.0946 & 0.0791 & [0.0848, 0.1269] & 0.0420 \\
\midrule
$\mathcal{T}_1$
& 0.5175 & 0.5131 & 0.1634 & [0.3907, 0.6414] & 0.2506 \\
$\mathcal{T}_1+\mathcal{T}_0$
& 0.5181 & 0.5032 & 0.1629 & [0.3873, 0.6512] & 0.2639 \\
$\mathcal{T}_1+2\mathcal{T}_0$
& 0.5324 & 0.5268 & 0.1672 & [0.3921, 0.6740] & 0.2819 \\
\midrule
$\mathcal{T}_2$
& 0.6612 & 0.6807 & 0.1344 & [0.5711, 0.7602] & 0.1891 \\
$\mathcal{T}_2+\mathcal{T}_0$
& 0.6664 & 0.6807 & 0.1343 & [0.5889, 0.7684] & 0.1796 \\
$\mathcal{T}_2+2\mathcal{T}_0$
& 0.6718 & 0.6933 & 0.1404 & [0.5747, 0.7766] & 0.2019 \\
\midrule
$\mathcal{T}_2+\mathcal{T}_1$
& 0.6974 & 0.7334 & 0.1282 & [0.6570, 0.7784] & 0.1214 \\
$\mathcal{T}_2+\mathcal{T}_1+\mathcal{T}_0$
& 0.6988 & 0.7346 & 0.1305 & [0.6448, 0.7828] & 0.1380 \\
$\mathcal{T}_2+\mathcal{T}_1+2\mathcal{T}_0$
& 0.7004 & 0.7409 & 0.1364 & [0.6536, 0.7902] & 0.1365 \\
\bottomrule
\end{tabular}
\end{table}

\textbf{Assessing Human Consistency.}
Holding each reference fixed, we assess whether predicted scores align with human severity judgments. For this evaluation, two PhD students independently annotated gap severity tiers, yielding the inter-annotator agreement summarized in Table~\ref{tab:human-agreement}. Agreement reaches 90.00\% at the gap level and 81.21\% at the transcript-pair level, with all kappa coefficients exceeding 0.83, and most disagreements occur between adjacent tiers, in line with weighted $\kappa$ increasing beyond the unweighted baseline.
\begin{table}[h]
\centering
\begin{minipage}[t]{0.48\textwidth}
\caption{Inter-annotator agreement. Gap-level agreement compares severity labels for each $g_i$, while pair-level agreement requires matching labels for all gaps associated with $h$.}
\label{tab:human-agreement}
\end{minipage}\hfill
\begin{minipage}[t]{0.48\textwidth}
\caption{Within-reference ranking agreement using final labels. $N$ counts strict-order comparisons; Agr. and Spearman $\rho_s$ are averaged over references.}
\label{tab:intrinsic-ranking}
\end{minipage}
\par
\begin{minipage}[t]{0.48\textwidth}
\centering
\begin{tabular*}{\linewidth}{@{\extracolsep{\fill}}lr@{}}
\toprule
Measure & Score \\
\midrule
$\kappa$ & 0.8396 \\
$\kappa_w^{\mathrm{lin}}$ & 0.8675 \\
$\kappa_w^{\mathrm{quad}}$ & 0.8924 \\
Gap-Level & 90.00\% \\
Pair-Level & 81.21\% \\
\bottomrule
\end{tabular*}
\end{minipage}\hfill
\begin{minipage}[t]{0.48\textwidth}
\centering
\begin{tabular*}{\linewidth}{@{\extracolsep{\fill}}lrrr@{}}
\toprule
Domain & $N$ & Agr. (\%) & Mean $\rho_s$ \\
\midrule
Culture & 495 & 88.48 & 0.8758 \\
Finance & 220 & 80.45 & 0.7614 \\
Services & 385 & 80.00 & 0.7078 \\
Civic & 275 & 83.64 & 0.8182 \\
General & 272 & 85.16 & 0.8062 \\
\bottomrule
\end{tabular*}
\end{minipage}
\end{table}
With human annotations directly adopted upon agreement and resolved via majority voting with initial model predictions in case of disagreement, we obtain 1{,}647 strict-order comparisons. \name achieves a mean pairwise agreement of 84.07\% and a mean $\rho_s$ of 0.8001. The median per-reference agreement of 85.45\% is close to the overall result, indicating similar ordering agreement for a typical reference. As shown in Table~\ref{tab:intrinsic-ranking}, all five evaluated domains attain mean pairwise agreement of at least 80.00\% and mean $\rho_s$ above 0.70. These results complement the cross-reference analysis by showing that most expected severity relations are also retained within individual references.

\section{Conclusion}
We introduced \name, a lightweight, deterministic metric for reference-grounded ASR situational distortion. Trained via CGD targets, a shared severity scale, and bounded recursive aggregation, it enables fast single-pass evaluation without runtime LLM calls. Across three English benchmarks and a controlled test set, \name\ demonstrates strong diagnostic value, severity ordering, and score consistency. Future work will scale data volume and multilingual coverage to study its scaling behavior.


\appendix
\section{SITAR Construction Details}
\subsection{Score Construction and Properties}
\label{app:score-construction}

Equation~\ref{eq:bounded-scalarization} combines a base score $\beta$ with a bounded refinement determined by gap counts. The base score records the most severe adjusted gap position, and the refinement records the remaining gaps according to their nominal tiers. This section gives the complete calculation, derives the parameter choice, and establishes the score's bounds, tier priority, and diminishing response to increasing counts.

\subsubsection{Base Score and Refinement Counts}

For a retained transcript pair, let $t_{\mathrm{rev}}^{(i)}$ and $t_{\mathrm{maj}}^{(i)}$ denote the reviewed and majority tiers of gap $g_i$, respectively. Following the annotation procedure in the main text, its adjusted severity position is
\begin{equation}
    \label{eq:app-gap-position}
    b_i=t_{\mathrm{rev}}^{(i)}
    +\frac{1}{3}\operatorname{sgn}\!\left(t_{\mathrm{maj}}^{(i)}-t_{\mathrm{rev}}^{(i)}\right),
    \qquad \beta=\max_i b_i,
\end{equation}
where $\operatorname{sgn}(0)=0$. Agreement leaves the nominal score unchanged, while disagreement shifts it by $1/3$ toward the majority tier. The reviewed tier remains the nominal assignment used for counting. The consistency filter in Eq.~\ref{eq:annotation-consistency} restricts retained majority and reviewed assignments to the same or adjacent tiers.

Explicitly, the possible $(t_{\mathrm{rev}}^{(i)},t_{\mathrm{maj}}^{(i)})$ pairs are $(0,0)$, $(0,1)$, $(1,0)$, $(1,1)$, $(1,2)$, $(2,1)$, and $(2,2)$, producing adjusted positions $0$, $1/3$, $2/3$, $1$, $4/3$, $5/3$, and $2$, respectively. These adjustments are computed for each gap before taking the maximum. A pair's base score therefore depends on the most severe adjusted gap, rather than an average of gap positions. The majority tier is obtained from the three annotations of that same gap; pairs without a per-gap majority have already been excluded by the annotation procedure.

We count each gap once using its reviewed tier,
\begin{equation}
    \label{eq:app-refinement-counts}
    c_k=\sum_i\mathbf{1}\!\left[t_{\mathrm{rev}}^{(i)}=k\right],
    \qquad k_{\max}=\max\{k:c_k>0\},
    \qquad \bar c_k=c_k-\mathbf{1}[k=k_{\max}].
\end{equation}
The gap that determines $\beta$ belongs to $\mathcal T_{k_{\max}}$, because even the lowest adjusted position at a higher nominal tier exceeds the highest position at the tier below it. Subtracting one at this tier prevents counting that gap again in the quantity refinement. All lower-tier gaps remain included. If several gaps attain the same maximum position, excluding any one of them gives the same refinement counts. The three original annotation rounds determine the majority assignments; they do not count as three separate observations of each gap.

Table~\ref{tab:score-base-intervals} lists the seven base positions and their score intervals. A pair with a single gap receives its base score directly. If no gaps are present, we define $S(r,h)=0$ separately, without taking a maximum over an empty set or entering the recursion.

\begin{table}[htbp]
    \centering
    \caption{Base positions and score intervals for finite gap counts. The highest nominal tier is determined by the reviewed assignments.}
    \label{tab:score-base-intervals}
    \begin{tabular}{lcc}
        \toprule
        Highest nominal tier & Base $\beta$ & Score interval \\
        \midrule
        $\mathcal T_0$ & $0$ & $[0,1/3)$ \\
        & $1/3$ & $[1/3,2/3)$ \\
        \midrule
        $\mathcal T_1$ & $2/3$ & $[2/3,1)$ \\
        & $1$ & $[1,4/3)$ \\
        & $4/3$ & $[4/3,5/3)$ \\
        \midrule
        $\mathcal T_2$ & $5/3$ & $[5/3,2)$ \\
        & $2$ & $[2,7/3)$ \\
        \bottomrule
    \end{tabular}
\end{table}

\subsubsection{Recursive Calculation}

For explicit computation of Eq.~\ref{eq:recursive-tier-aggregation}, initialize $u_{-1}=0$ and process every nominal tier up to $k_{\max}$ in ascending order,
\begin{equation}
    \label{eq:app-score-recursion}
    u_k=1-\rho^{\bar c_k+u_{k-1}},
    \qquad k=0,\ldots,k_{\max},
    \qquad S(r,h)=\beta+\frac{u_{k_{\max}}}{3}.
\end{equation}
An intermediate tier with zero gaps is still evaluated. Its update becomes $u_k=1-\rho^{u_{k-1}}$, preserving the tier through which lower-level contributions pass. Skipping such a tier would change the score.

For convenience, define $U_\rho(c,u)=1-\rho^{c+u}$ and write $R=u_{k_{\max}}$ for the final refinement. The three possible highest tiers give
\begin{equation}
    \label{eq:app-recursion-by-tier}
    R=
    \begin{cases}
        U_\rho(c_0-1,0), & k_{\max}=0,\\
        U_\rho\!\left(c_1-1,U_\rho(c_0,0)\right), & k_{\max}=1,\\
        U_\rho\!\left(c_2-1,U_\rho\!\left(c_1,U_\rho(c_0,0)\right)\right), & k_{\max}=2.
    \end{cases}
\end{equation}
Only the highest occupied tier has its count reduced by one. A missing lower tier contributes zero new gaps but still applies its recursive update. Since $U_\rho(0,0)=0$, leading empty tiers do not change an otherwise single-tier calculation. All updates use unrounded intermediate values; the numerical examples below are rounded only for presentation.

\subsubsection{Parameter Choice}

\paragraph{Refinement width.}
Suppose the score is written as $S=\beta+\lambda R$. Adjacent base positions are separated by $1/3$, while the recursive refinement satisfies $0\leq R<1$ and approaches one as the highest-tier count increases. Any $0<\lambda\leq1/3$ preserves the interval boundaries. We use $\lambda=1/3$ to make the full interval available for quantity refinement. A smaller value would leave part of each interval unused even in the infinite-count limit, and a larger value would allow sufficiently many gaps to cross the next base position. The selected width is thus fixed by the base-score grid rather than tuned during regressor training.

\paragraph{Saturation rate.}
To set $\rho$, consider $n$ gaps all assigned to the same nominal tier. After the first gap is accounted for by $\beta$, the refinement simplifies to
\begin{equation}
    \label{eq:app-same-tier-refinement}
    R_n=1-\rho^{n-1},
    \qquad S_n=\beta+\frac{1-\rho^{n-1}}{3}.
\end{equation}
We choose seven same-tier gaps as a reference count and require them to occupy 90\% of the refinement interval. Thus,
\begin{equation}
    \label{eq:app-rho-choice}
    R_7=0.9
    \quad\Longrightarrow\quad \rho^6=0.1
    \quad\Longrightarrow\quad \rho=10^{-1/6}\approx0.681292.
\end{equation}
More generally, a reference count $N_{\mathrm{ref}}>1$ and utilization $0<\eta<1$ give $\rho=(1-\eta)^{1/(N_{\mathrm{ref}}-1)}$. Seven is a calibration point, not a cap on the number of gaps. Table~\ref{tab:same-tier-refinement} shows how the refinement continues to grow beyond this count.

\begin{table}[htbp]
    \centering
    \caption{Quantity refinement for $n$ gaps at the same nominal tier, with $\rho=10^{-1/6}$. Values are rounded to six decimal places.}
    \label{tab:same-tier-refinement}
    \begin{tabular}{rrr}
        \toprule
        Gaps $n$ & Refinement $R_n$ & Score increment $R_n/3$ \\
        \midrule
        1 & 0.000000 & 0.000000 \\
        2 & 0.318708 & 0.106236 \\
        3 & 0.535841 & 0.178614 \\
        4 & 0.683772 & 0.227924 \\
        5 & 0.784557 & 0.261519 \\
        6 & 0.853220 & 0.284407 \\
        7 & 0.900000 & 0.300000 \\
        8 & 0.931871 & 0.310624 \\
        9 & 0.953584 & 0.317861 \\
        10 & 0.968377 & 0.322792 \\
        11 & 0.978456 & 0.326152 \\
        \bottomrule
    \end{tabular}
\end{table}

\subsubsection{Properties of the Score}

\paragraph{Bounded refinement.}
For $0<\rho<1$, a finite nonnegative integer $c$, and $0\leq u<1$, the update $1-\rho^{c+u}$ lies in $[0,1)$. Starting from $u_{-1}=0$, induction therefore gives $0\leq u_k<1$ at every tier. Consequently,
\begin{equation}
    \label{eq:app-score-bounds}
    \beta\leq S(r,h)<\beta+\frac13\leq\frac73.
\end{equation}
The base intervals are disjoint. Combining the intervals associated with each highest nominal tier gives $[0,2/3)$ for $\mathcal T_0$, $[2/3,5/3)$ for $\mathcal T_1$, and $[5/3,7/3)$ for $\mathcal T_2$. Quantity refinement therefore cannot move a pair into a higher base interval or nominal tier.

\paragraph{Priority of higher tiers.}
The update $U_\rho(c,u)$ is strictly increasing in both arguments. More strongly, for every $0\leq u<1$,
\begin{equation}
    \label{eq:app-tier-priority}
    U_\rho(c+1,0)-U_\rho(c,u)
    =\rho^c\left(\rho^u-\rho\right)>0.
\end{equation}
An additional gap at the current tier thus outweighs any finite configuration at lower tiers. For two count configurations with the same $\beta$, consider the highest tier at which their counts differ. At that tier, the configuration with the larger integer count has the larger recursive result, even if its lower-tier contribution is zero and the other configuration's lower-tier contribution is arbitrarily close to one. Counts at all higher tiers agree, so their strictly increasing updates preserve this ordering. The refinement therefore orders distinct count configurations lexicographically from the highest tier downward. At a fixed highest nominal tier, this also means that distinct count configurations cannot have identical refinements in exact arithmetic.

\paragraph{Increasing counts with diminishing increments.}
Fix $\beta$ and $k_{\max}$. To study sensitivity to a count, extend the update to nonnegative real-valued counts and let $a=-\log\rho>0$. Its first and second derivatives with respect to the current count are
\begin{equation}
    \label{eq:app-count-derivatives}
    \frac{\partial U_\rho(c,u)}{\partial c}=a\rho^{c+u}>0,
    \qquad
    \frac{\partial^2 U_\rho(c,u)}{\partial c^2}=-a^2\rho^{c+u}<0.
\end{equation}
The same derivatives apply with respect to $u$. Holding all other counts fixed, each subsequent tier applies another strictly increasing, concave function. Their composition is therefore strictly increasing and concave in the count being varied. Restricting back to integer counts, adding a gap at any tier up to $k_{\max}$ increases the score, while successive additions at that same tier have decreasing effects. In the single-tier case, Eq.~\ref{eq:app-same-tier-refinement} gives the increment explicitly,
\begin{equation}
    \label{eq:app-diminishing-increment}
    S_{n+1}-S_n=\frac{(1-\rho)\rho^{n-1}}{3}>0.
\end{equation}
Each successive increment is multiplied by $\rho$, so the score approaches $\beta+1/3$ without reaching it at any finite count. For benign speech disfluencies assigned to $\mathcal T_0$, this refinement records their accumulation while retaining the situation-preserving nominal tier.

\paragraph{A bound for a fixed total count.}
Let $n=\sum_k c_k\geq1$ and let $L=\sum_{k<k_{\max}}c_k$ count all lower-tier gaps. If $L=0$, then $R=1-\rho^{n-1}$. If $L\geq1$, the lower-tier recursive result satisfies $u_{k_{\max}-1}<1\leq L$, giving
\begin{equation}
    \label{eq:app-total-count-bound}
    R=1-\rho^{c_{k_{\max}}-1+u_{k_{\max}-1}}
    <1-\rho^{c_{k_{\max}}-1+L}
    =1-\rho^{n-1}.
\end{equation}
Thus $R\leq1-\rho^{n-1}$, with equality when all gaps occupy the highest tier. For at most $N$ gaps, the distance between the largest attainable score in one base interval and the next base position is at least $\rho^{N-1}/3$. With $N=7$, this gives $R\leq0.9$, $S\leq2.3$, and a separation of at least $1/30$. The universal bound $S<7/3$ remains valid when the count exceeds seven.

\subsubsection{Worked Examples}

\paragraph{A complete mixed-tier calculation.}
Consider the illustrative annotations in Table~\ref{tab:score-worked-annotations}. The reviewed assignments contain two $\mathcal T_0$ gaps, one $\mathcal T_1$ gap, and two $\mathcal T_2$ gaps. For $g_4$, the majority tier is one but the reviewed tier is two, so its adjusted position is $5/3$. Gap $g_3$ has adjusted position two and sets $\beta=2$. The counts are $(c_0,c_1,c_2)=(2,1,2)$, and excluding one highest-tier gap gives $(\bar c_0,\bar c_1,\bar c_2)=(2,1,1)$.

\begin{table}[htbp]
    \centering
    \caption{Illustrative gap annotations for the mixed-tier calculation. The three original labels determine the majority tier, and the reviewed tier determines the nominal count.}
    \label{tab:score-worked-annotations}
    \begin{tabular}{lcccc}
        \toprule
        Gap & Original labels & Majority tier & Reviewed tier & Position $b_i$ \\
        \midrule
        $g_0$ & $(0,0,0)$ & 0 & 0 & $0$ \\
        $g_1$ & $(0,0,1)$ & 0 & 0 & $0$ \\
        $g_2$ & $(1,1,0)$ & 1 & 1 & $1$ \\
        $g_3$ & $(2,2,1)$ & 2 & 2 & $2$ \\
        $g_4$ & $(1,1,2)$ & 1 & 2 & $5/3$ \\
        \bottomrule
    \end{tabular}
\end{table}

Applying Eq.~\ref{eq:app-score-recursion} gives
\begin{equation}
    \label{eq:app-mixed-tier-example}
    \begin{aligned}
        u_0&=1-\rho^2\approx0.535841,\\
        u_1&=1-\rho^{1+u_0}\approx0.445340,\\
        u_2&=1-\rho^{1+u_1}\approx0.425738,\\
        S(r,h)&=2+u_2/3\approx2.141913.
    \end{aligned}
\end{equation}
If the original labels of $g_3$ are instead $(1,1,2)$ while its reviewed tier remains two, its adjusted position becomes $5/3$. Both highest-tier gaps then have that position, giving $\beta=5/3$. The reviewed counts and quantity refinement remain unchanged, and the final score becomes $1.808579$. The base positions differ by $1/3$, and the final scores retain exactly this difference.

\paragraph{Higher-tier count versus lower-tier accumulation.}
For a direct comparison of tier priority, fix $\beta=1$. One $\mathcal T_1$ gap with six $\mathcal T_0$ gaps receives
\[
    S_A=1+\frac{1-\rho^{\,1-\rho^6}}{3}\approx1.097351,
\]
whereas two $\mathcal T_1$ gaps with no lower-tier gaps receive
\[
    S_B=1+\frac{1-\rho}{3}\approx1.106236.
\]
Thus $S_A<S_B$, as required by Eq.~\ref{eq:app-tier-priority}.

\paragraph{An empty intermediate tier.}
For $(c_0,c_1,c_2)=(2,0,1)$ and $\beta=2$, the refinement counts are $(2,0,0)$. The successive values are
\[
    u_0=1-\rho^2\approx0.535841,
    \qquad u_1=1-\rho^{u_0}\approx0.185871,
    \qquad u_2=1-\rho^{u_1}\approx0.068846.
\]
The resulting score is $2.022949$. Skipping the empty $\mathcal T_1$ tier would instead give $2.061957$, incorrectly treating the $\mathcal T_0$ contribution as if it were immediately below $\mathcal T_2$.

\paragraph{Situation-preserving disfluencies.}
When all reviewed gaps belong to $\mathcal T_0$ and their majority assignments agree, $\beta=0$. One such gap receives score zero, two receive $0.106236$, and seven receive $0.300000$. Further gaps increase the score toward $1/3$ while remaining strictly below it. These positive refinements record the accumulation of expression-level differences, including benign speech disfluencies, without changing the nominal assessment that the situation is preserved. A zero score is therefore compatible with either no gap or a single unadjusted $\mathcal T_0$ gap.

\subsubsection{Finite-Count Verification}

We enumerate all nonempty count configurations with at most seven gaps,
\[
    \sum_{n=1}^{7}\binom{n+2}{2}=119.
\]
These comprise 7 configurations with highest tier $\mathcal T_0$, 28 with $\mathcal T_1$, and 84 with $\mathcal T_2$. All satisfy $0\leq R\leq0.9$. Sorting the configurations within each highest tier by $(c_{k_{\max}},\ldots,c_0)$ produces strictly increasing refinements, with 7, 28, and 84 distinct values, respectively. The largest refinement is $0.9$, attained by seven gaps at the same tier. This finite enumeration checks the calculations at the chosen calibration count; the bounds and ordering for arbitrary finite counts follow from the preceding proofs. Distinctness refers to the unrounded mathematical scores, since rounding can merge nearby values.

\subsection{Training Data Construction}
\label{app:training-data-construction}

We primarily draw on AppTek Call-Center Dialogues~\citep{beck26_interspeech}, a newly collected conversational corpus with professional verbatim transcripts that preserve disfluencies, repetitions, and self-repairs. Its 1,746 single-channel recordings are segmented into 31,354 short clips spanning 14 accent configurations and 16 service domains. To extend this conversational coverage beyond customer-service interactions, we supplement the clips with 3,643 English recordings from FLEURS~\citep{10023141}, which provides read speech based on Wikipedia passages. Together, the two sources yield 34,997 speech segments at 16~kHz, with their audio and reference-transcript length distributions summarized in Table~\ref{tab:candidate-speech-pool}.
\begin{table}[h]
    \centering
    \caption{Statistics of candidate audio and reference transcripts across AppTek and FLEURS datasets.}
    \label{tab:candidate-speech-pool}
    \begin{tabular}{llrrrrrr}
        \toprule
        Type & Source & N & Mean & P25 & P50 & P75 & P95 \\
        \midrule
        \multirow{2}{*}{Audio}
        & AppTek\_Clip & 31,354 & 10.77~s & 3.14~s & 7.66~s & 15.90~s & 31.79~s \\
        & FLEURS\_en   & 3,643  & 10.19~s & 7.68~s & 9.66~s & 11.88~s & 16.90~s \\
        \midrule
        \multirow{2}{*}{Transcript}
        & AppTek\_Clip & 31,354 & 24.26 & 8  & 18 & 34 & 70 \\
        & FLEURS\_en   & 3,643  & 21.21 & 16 & 20 & 25 & 34 \\
        \bottomrule
    \end{tabular}
\end{table}

\begin{figure}[h]
    \centering
    \includegraphics[width=0.82\linewidth]{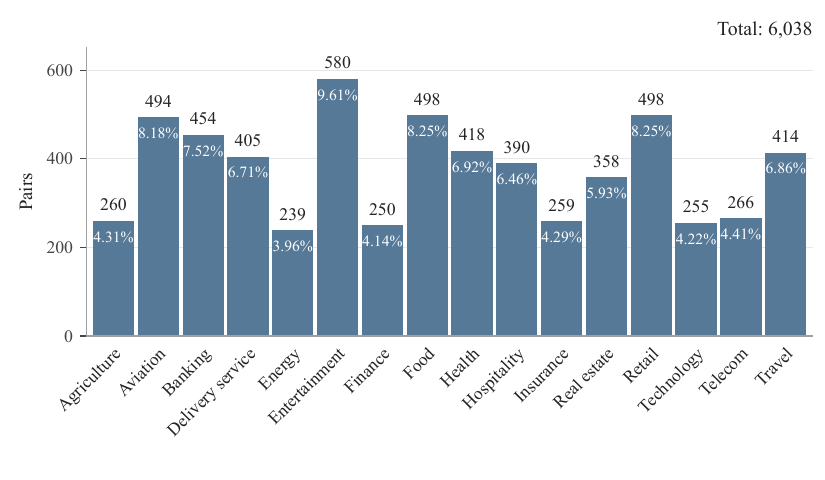}
    \caption{Domain distribution of the 6,038 selected AppTek pairs.}
    \label{fig:apptek-domain-distribution}
\end{figure}

This speech pool supplies the reference--hypothesis pairs used to construct supervision for \name. We transcribe the segments using Qwen3-ASR-1.7B, Granite-Speech-4.1, Whisper-large-v3, and Parakeet-v3, pairing each output with the corresponding reference. Applying multiple recognizers to the same speech allows a shared reference to be associated with different naturally occurring transcription discrepancies, with each recognizer's output forming a separate candidate pair.
\begin{table}[h]
    \centering
    \caption{Selected reference--hypothesis pairs by source, locale, and ASR system.}
    \label{tab:candidate-pair-distribution}
    \begin{tabular}{llrrrrr}
        \toprule
        Source & Locale & Qwen3 & Granite & Whisper & Parakeet & Total \\
        \midrule
        AppTek\_Clip & en-AU & 188 & 190 & 167 & 155 & 700 \\
        & en-CA          & 151 & 157 & 150 & 187 & 645 \\
        & en-CN          & 185 & 151 & 160 & 177 & 673 \\
        & en-GB          & 162 & 167 & 177 & 152 & 658 \\
        & en-IE          & 184 & 165 & 171 & 151 & 671 \\
        & en-MX          & 176 & 164 & 167 & 155 & 662 \\
        & en-SG          & 164 & 158 & 159 & 163 & 644 \\
        & en-US\_General & 178 & 156 & 163 & 164 & 661 \\
        & en-ZA          & 187 & 184 & 171 & 182 & 724 \\
        \midrule
        AppTek subtotal & & 1,575 & 1,492 & 1,485 & 1,486 & 6,038 \\
        FLEURS\_en & en\_us & 500 & 500 & 500 & 500 & 2,000 \\
        \midrule
        Total & & 2,075 & 1,992 & 1,985 & 1,986 & 8,038 \\
        \bottomrule
    \end{tabular}
\end{table}
From these candidates, we select 8,038 pairs for annotation, restricting audio duration to 10--33~s and balancing recognizer contributions. For AppTek, sampling jointly balances ASR systems and accents, yielding 6,038 pairs across nine accents while covering all 16 service domains. This conversational subset is complemented by 2,000 FLEURS pairs, split equally across the four recognizers. Table~\ref{tab:candidate-pair-distribution} shows candidate distributions across sources, recognizers, and accents, while Figure~\ref{fig:apptek-domain-distribution} details domain coverage for AppTek.

We annotate the selected pairs using DeepSeek-V4-Pro across three independent rounds. We crafted a dedicated prompt to guide the model, ensuring strict adherence to a multilevel rubric grounded in the OECD results-chain framework. This initial rubric preserves finer distinctions among levels of consequence than the three-tier scheme ultimately adopted by \name. In practice, however, the highest consequence levels are represented by very few examples. These low-frequency categories carry greater surprisal in an information-theoretic sense and, in our annotations, correspond to more severe situational consequences. Their scarcity nevertheless provides insufficient support for reliable analysis and model training at this granularity. We therefore consolidate the higher consequence tiers into $\mathcal{T}_2$, yielding the final three-tier scheme used throughout the paper.
The annotations subsequently undergo review by a committee of four agents using GPT-5.6-Sol-Max in ultra mode. The committee reassesses each gap's severity in the context of its reference and supplies the reviewed tier assignments used to construct the scalar supervision targets.

We split the annotated pairs into training, validation, and test sets using proportions of 75\%, 10\%, and 15\%, yielding 6,436, 856, and 1,292 pairs, respectively. These pairs come from errors that existing ASR systems happen to produce, so we cannot directly control their types or frequencies. We need severe-error examples to train the metric, but ASR systems are optimized to reduce transcription errors. Collecting such examples therefore requires searching through a large number of transcriptions, and some score intervals remain sparsely represented in the resulting training set.

We augment the training data by constructing variants of pairs containing at least three gaps. For each pair, we selectively correct a subset of the gaps while preserving the highest severity level, then recompute the supervision score from the remaining gaps. We preferentially select variants whose scores fall in sparsely populated intervals. This adds 3,763 unique pairs, increasing the training set from 6,436 to 10,199 pairs. Together with the validation and test sets, the dataset contains 12,347 unique pairs.

\begin{figure}[t]
    \centering
    \includegraphics[width=\linewidth]{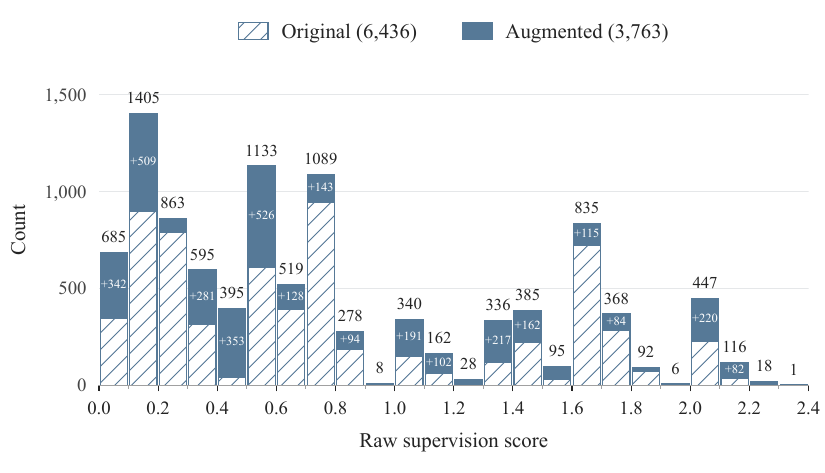}
    \caption{Distribution of unique training pairs across supervision score bins (width 0.1). Hatched and solid blue regions denote original and added pairs, respectively. Values above bars show total counts; labels inside (`+' prefixed) indicate added counts.}
    \label{fig:training-score-augmentation}
\end{figure}

Figure~\ref{fig:training-score-augmentation} shows the original and added training pairs in score intervals of width 0.1. For intervals that remain sparse after augmentation, we upsample the training pairs by duplicating each pair once.

Although annotation starts from three severity tiers with nominal scores of 0, 1, and 2, directional boundary adjustments and the bounded recursive aggregation of gap counts yield finer score distinctions. Different combinations, such as $2\mathcal{T}_2+\mathcal{T}_1$ and $\mathcal{T}_2+3\mathcal{T}_1+\mathcal{T}_0$, can therefore receive different scores despite sharing the same highest severity tier.

\begin{figure}[t]
    \centering
    \includegraphics[width=\linewidth]{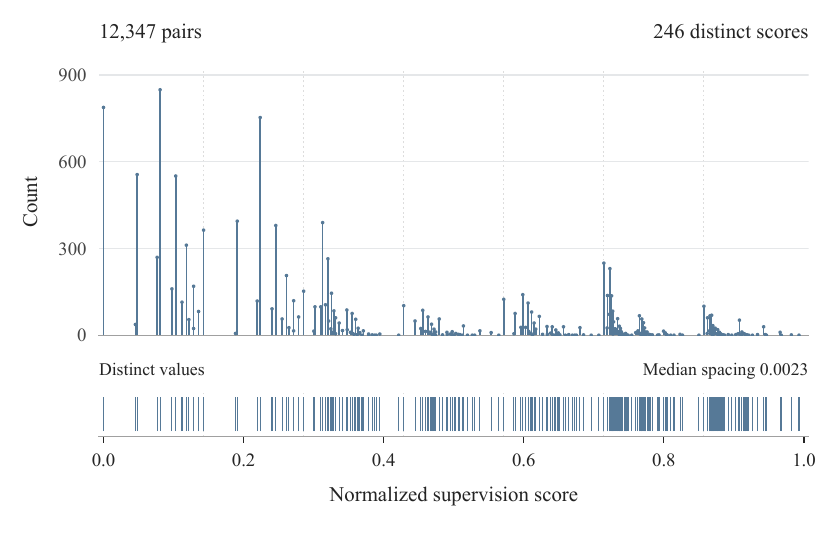}
    \caption{Normalized supervision scores across all 12,347 pairs. The upper panel shows the sample count at each score, and the lower panel marks the positions of the 246 distinct values. Distinct values are counted by rounding raw scores to four decimal places before normalization. Dashed lines mark the normalized base-score interval boundaries.}
    \label{fig:dataset-score-distribution}
\end{figure}

We further examine the supervision scores across all 12,347 pairs. When rounded to four decimal places, the raw scores contain 246 distinct values within the theoretical range of $[0,7/3)$. Figure~\ref{fig:dataset-score-distribution} shows their distribution after dividing the raw scores by $7/3$ to normalize them to $[0,1)$. On this normalized scale, the observed values range from 0 to 0.9929, with a median spacing of 0.0023 between adjacent distinct values. These values provide a fine-grained, pseudo-continuous supervision scale that captures differences in both error severity and gap composition.

\clearpage

\subsection{Scorer Construction}
\label{app:scorer-construction}

This section details the implementation of the \name scorer and the experiments supporting its configuration. The objective is to learn the scalar supervision targets described in Section~\ref{sec:methodology} from a reference--hypothesis pair. Accordingly, the regression experiments below assess how well the scorer reproduces these targets; the metric-validity evaluations in the main text examine its behavior against complementary criteria.

\subsubsection{Architecture and Optimization}
\label{app:scorer-implementation}

We jointly encode the reference and hypothesis using \texttt{microsoft/deberta-v3-large}. The tokenizer constructs a sentence-pair input, with dynamic padding and truncation to a maximum of 512 tokens, including special tokens. The final configuration uses the original transcripts without explicit gap markers. Last-layer token representations are passed through a single bidirectional LSTM, whose hidden size in each direction is half the encoder hidden size. We then average its outputs over non-padding positions to obtain a fixed-dimensional pair representation. Sequences are packed before the LSTM so that padding does not enter the recurrent computation.

The pooled representation is passed through dropout with probability 0.1 and two separate linear heads: a scalar regression head and an auxiliary ordinal head. Only the scalar head supplies the \name prediction. All encoder, pooling, and head parameters are optimized jointly in the final model. Table~\ref{tab:scorer-training-settings} lists the training configuration.

\begin{table}[!htbp]
    \centering
    \caption{Configuration of the final \name scorer. Learning rates distinguish the pretrained encoder from the LSTM and prediction heads.}
    \label{tab:scorer-training-settings}
    \begin{tabular}{@{}p{0.43\linewidth}p{0.53\linewidth}@{}}
        \toprule
        Component & Setting \\
        \midrule
        Encoder & DeBERTa-v3-large; full fine-tuning \\
        Input and maximum length & Original reference--hypothesis pair; 512 tokens \\
        Pooling & One-layer bidirectional LSTM, followed by masked mean \\
        Prediction heads & Linear scalar head and six-logit ordinal head \\
        Head dropout & 0.1 \\
        Loss weights $(\lambda_{\mathrm{mse}},\lambda_{\mathrm{boundary}},\lambda_{\mathrm{ordinal}})$ & $(1.0,0.2,0.5)$ \\
        Ordinal thresholds & $\{1/3,2/3,1,4/3,5/3,2\}$ \\
        Boundary refinement & $\rho=10^{-1/6}$; $q=2$ \\
        Optimizer & AdamW \\
        Encoder / pooling and head learning rates & $6\!\times\!10^{-6}$ / $10^{-4}$ \\
        Learning-rate schedule & 5\% linear warmup, then cosine decay \\
        Batch size / gradient accumulation & 8 / 4 steps \\
        Maximum epochs / early-stopping patience & 10 / 3 validation checks \\
        Gradient clipping / precision & Global norm 1.0 / FP32 \\
        Checkpoint criterion & Lowest validation MSE within each run \\
        Configuration and final seed selection & Validation MSE \\
        Random seeds for final retraining & 42, 43, 44, 45, 46 \\
        \bottomrule
    \end{tabular}
\end{table}

AdamW uses separate learning rates for encoder parameters and newly trained pooling and head parameters. Biases, normalization parameters, and other one-dimensional parameters are excluded from weight decay. Linear warmup occupies the first 5\% of planned optimizer updates; cosine decay subsequently reduces the learning rate to 10\% of its peak at the final planned update. Validation MSE controls checkpoint retention and early stopping. All experiments use the same data partition, comprising 10,199 unique training pairs, 856 validation pairs, and 1,292 test pairs, as described in Appendix~\ref{app:training-data-construction}.

\subsubsection{Training Objective}
\label{app:scorer-objective}

Let $y_n$ be the scalar target of pair $(r_n,h_n)$, $s_n=f_\theta(r_n,h_n)$ its regression output, and $z_{nj}$ the ordinal logit associated with threshold $\tau_j$. For a minibatch of $N$ pairs, the final objective is
\begin{equation}
    \label{eq:app-training-objective}
    \mathcal L
    =\mathcal L_{\mathrm{mse}}
    +0.2\mathcal L_{\mathrm{boundary}}
    +0.5\mathcal L_{\mathrm{ordinal}}.
\end{equation}
The three terms supervise numerical accuracy, interval membership, and cumulative threshold prediction, respectively:
\begin{equation}
    \label{eq:app-loss-components}
    \begin{aligned}
        \mathcal L_{\mathrm{mse}}
        &=\frac{1}{N}\sum_{n=1}^{N}(s_n-y_n)^2,\\
        \mathcal L_{\mathrm{boundary}}
        &=\frac{1}{N}\sum_{n=1}^{N}
        \left([\ell_n-s_n]_+^2+[s_n-u_n]_+^2\right),\\
        \mathcal L_{\mathrm{ordinal}}
        &=\frac{1}{NJ}\sum_{n=1}^{N}\sum_{j=1}^{J}
        \operatorname{BCEWithLogits}\!\left(z_{nj},\mathbf 1[y_n\geq\tau_j]\right),
    \end{aligned}
\end{equation}
where $[x]_+=\max(x,0)$, $J=6$, and $(\tau_1,\ldots,\tau_J)=(1/3,2/3,1,4/3,5/3,2)$. The ordinal targets are derived entirely from $y_n$, requiring no additional annotations. All thresholds and positive classes receive unit weight in the final configuration.

The bounds $\ell_n$ and $u_n$ are the adjacent points of a fixed boundary grid containing $y_n$. To construct this grid, let $(a_0,\ldots,a_{J+1})=(0,\tau_1,\ldots,\tau_J,7/3)$. Each coarse interval is subdivided using
\begin{equation}
    \label{eq:app-boundary-grid}
    \mathcal B_q
    =\bigcup_{j=0}^{J}
    \left\{a_j+(1-\rho^k)(a_{j+1}-a_j):k=0,\ldots,q-1\right\}
    \cup\{7/3\}.
\end{equation}
We use $\rho=10^{-1/6}$ and $q=2$, giving 14 fine intervals. Intervals are left-closed and right-open, except that the last includes $7/3$; comparisons use a tolerance of $10^{-6}$ for rounded labels. Boundary loss is zero for predictions within the target interval and grows quadratically outside it. This grid is a training constraint on the scalar prediction; constructing it does not require gap counts at inference time.

\subsubsection{Design Ablations}
\label{app:scorer-ablations}

We examine pooling, auxiliary losses, input representation, and encoder adaptation while retaining the same data partition. Checkpoints within each run, model configurations, and the final random seed are selected using validation MSE. The regression results in this and the following subsections are reported on the 1,292-pair test split against the constructed scalar targets, using raw, unclipped predictions. Lower MSE and MAE indicate smaller numerical errors, while higher Pearson correlation $r$ and Spearman correlation $\rho_s$ indicate closer agreement with the targets. Multi-seed results are means and sample standard deviations.

\paragraph{Pooling.}
We first compare five pooling methods under MSE-only training with seed 42. In addition to the last-layer first-token representation (CLS) and masked mean, we consider concatenation and a learned weighted mixture of the mean-pooled last four layers, as well as LSTM pooling. We select LSTM pooling by validation MSE for the subsequent regression experiments. On the test split, Table~\ref{tab:scorer-pooling} shows that LSTM pooling achieves the lowest MSE and MAE, whereas CLS yields the highest Spearman correlation.

\begin{table}[!htbp]
    \centering
    \caption{Pooling comparison with MSE-only training and seed 42. Bold values indicate the best result in each column.}
    \label{tab:scorer-pooling}
    \begin{tabular}{lrrrr}
        \toprule
        Pooling & MSE $\downarrow$ & MAE $\downarrow$ & Pearson $r$ $\uparrow$ & Spearman $\rho_s$ $\uparrow$ \\
        \midrule
        CLS & 0.138204 & 0.261133 & 0.810322 & \textbf{0.813424} \\
        Mean & 0.143248 & 0.277111 & 0.795377 & 0.793949 \\
        Last-four-layer concatenation & 0.150199 & 0.275274 & 0.794139 & 0.789104 \\
        Learned layer mixture & 0.147843 & 0.267151 & 0.798982 & 0.796896 \\
        LSTM & \textbf{0.129790} & \textbf{0.255395} & \textbf{0.812736} & 0.807285 \\
        \bottomrule
    \end{tabular}
\end{table}

\paragraph{Auxiliary losses.}
An initial comparison of MSE alone and its combinations with boundary and ordinal losses under CLS and LSTM pooling yielded a lowest MSE of 0.128095 for LSTM with boundary loss. We then searched a $9\times9$ grid with $\lambda_{\mathrm{boundary}},\lambda_{\mathrm{ordinal}}\in\{0.1,0.2,\ldots,0.9\}$, fixing $\lambda_{\mathrm{mse}}=1$ and seed 42. Neither weight exhibited a monotonic relationship with performance: the lowest MSE occurred at $(0.3,0.9)$, whereas the highest Spearman correlation occurred at $(0.2,0.3)$.

To examine sensitivity to initialization, we evaluated eight loss configurations using seeds 42--46 (Table~\ref{tab:scorer-loss-ablation}). Validation MSE is used to select $(\lambda_{\mathrm{boundary}},\lambda_{\mathrm{ordinal}})=(0.2,0.5)$ for the remaining ablations and final retraining. On the test split, this combination gives the lowest mean MSE, reducing it from 0.134565 to 0.127040 relative to MSE-only training, a relative reduction of 5.59\%. It also gives the highest mean Spearman correlation among these eight configurations.

\begin{table}[t]
    \centering
    \caption{Loss-weight comparison with LSTM pooling and full encoder fine-tuning. Results are mean $\pm$ sample standard deviation over five seeds; $\lambda_{\mathrm{mse}}=1$ throughout. Bold marks the best means.}
    \label{tab:scorer-loss-ablation}
    \begin{tabular}{ccrr}
        \toprule
        $\lambda_{\mathrm{boundary}}$ & $\lambda_{\mathrm{ordinal}}$ & MSE $\downarrow$ & Spearman $\rho_s$ $\uparrow$ \\
        \midrule
        0.0 & 0.0 & $0.134565\pm0.004427$ & $0.802046\pm0.002034$ \\
        0.0 & 0.1 & $0.131830\pm0.001990$ & $0.802677\pm0.002407$ \\
        0.0 & 0.9 & $0.131293\pm0.001865$ & $0.808193\pm0.003262$ \\
        0.3 & 0.0 & $0.133434\pm0.002802$ & $0.803971\pm0.005725$ \\
        0.3 & 0.9 & $0.130475\pm0.002781$ & $0.806029\pm0.006550$ \\
        0.2 & 0.5 & $\mathbf{0.127040}\pm0.004466$ & $\mathbf{0.808929}\pm0.005065$ \\
        0.4 & 0.1 & $0.130092\pm0.003100$ & $0.804058\pm0.002717$ \\
        0.4 & 0.8 & $0.130238\pm0.003917$ & $0.807426\pm0.003030$ \\
        \bottomrule
    \end{tabular}
\end{table}

\begin{table}[h]
    \centering
    \caption{Input and encoder-adaptation ablations. Each row reports five-seed mean $\pm$ sample standard deviation with LSTM pooling and loss weights $(1,0.2,0.5)$. The original-input and full-fine-tuning rows reuse the same baseline runs.}
    \label{tab:scorer-input-freezing}
    \begin{tabular}{lrr}
        \toprule
        Configuration & MSE $\downarrow$ & Spearman $\rho_s$ $\uparrow$ \\
        \midrule
        \multicolumn{3}{l}{\textit{Input representation; encoder fully trainable}} \\
        Original reference--hypothesis pair & $\mathbf{0.127040}\pm0.004466$ & $\mathbf{0.808929}\pm0.005065$ \\
        Inline gap markers & $0.136698\pm0.002313$ & $0.802945\pm0.003336$ \\
        Appended gap spans & $0.140356\pm0.005461$ & $0.798801\pm0.003963$ \\
        \midrule
        \multicolumn{3}{l}{\textit{Encoder adaptation; original transcript input}} \\
        Frozen encoder & $0.161915\pm0.002660$ & $0.764525\pm0.004795$ \\
        Last 4 layers & $0.141614\pm0.001690$ & $0.794641\pm0.003432$ \\
        Last 8 layers & $0.138390\pm0.004855$ & $0.803412\pm0.002441$ \\
        Last 12 layers & $0.137513\pm0.004675$ & $0.799948\pm0.004617$ \\
        Last 16 layers & $0.136044\pm0.003796$ & $0.803115\pm0.002518$ \\
        Last 20 layers & $0.134422\pm0.002989$ & $0.802073\pm0.005120$ \\
        Last 24 layers & $0.128223\pm0.003758$ & $0.807644\pm0.001487$ \\
        Full fine-tuning & $\mathbf{0.127040}\pm0.004466$ & $\mathbf{0.808929}\pm0.005065$ \\
        \bottomrule
    \end{tabular}
\end{table}

\paragraph{Input representation and encoder adaptation.}
We compare the original transcript pair with two ways of exposing CGD gaps to the encoder: inserting markers around gaps at their original positions, or appending marked gap spans to each transcript. All three use LSTM pooling and the same loss weights. Validation MSE is used to select the original transcript pair as input, while CGD is used during offline target construction. As shown in Table~\ref{tab:scorer-input-freezing}, neither form of explicit gap input improves the five-seed mean test MSE or Spearman correlation.

We also compare full encoder fine-tuning with a frozen encoder and with training only its last $K$ Transformer layers. The pooling and prediction heads remain trainable in every condition. Increasing the number of trainable encoder layers generally reduces MSE, with $K=24$ approaching full fine-tuning. The $K$-layer policy keeps embeddings and other encoder parameters outside the selected layers frozen, so it differs from full fine-tuning even when all Transformer layers are trainable. Full fine-tuning is selected using validation MSE and also gives the lowest mean test MSE in this comparison.

\subsubsection{Hyperparameter Sensitivity}
\label{app:scorer-sensitivity}

\paragraph{Learning rates.}
We vary the encoder learning rate over $\{3\times10^{-6},6\times10^{-6},10^{-5},2\times10^{-5}\}$ and the pooling/head learning rate over $\{5\times10^{-5},10^{-4},2\times10^{-4},4\times10^{-4}\}$, using seeds 42 and 43. Validation MSE is used to select $(6\times10^{-6},10^{-4})$ for the final configuration. Table~\ref{tab:scorer-learning-rates} reports mean test MSE for all 16 combinations. The selected pair gives the lowest mean test MSE, whereas $(6\times10^{-6},2\times10^{-4})$ gives the highest mean test Spearman correlation (0.812878 versus 0.811107). The latter also yields the lowest single-run test MSE in the search (0.121435).

\begin{table}[h]
    \centering
    \caption{Learning-rate sensitivity: mean MSE over seeds 42 and 43. Columns specify the learning rate for pooling and prediction heads. The bold entry is the lowest mean.}
    \label{tab:scorer-learning-rates}
    \begin{tabular}{lrrrr}
        \toprule
        Encoder learning rate & $5\times10^{-5}$ & $10^{-4}$ & $2\times10^{-4}$ & $4\times10^{-4}$ \\
        \midrule
        $3\times10^{-6}$ & 0.134568 & 0.133172 & 0.133939 & 0.134907 \\
        $6\times10^{-6}$ & 0.128956 & \textbf{0.123872} & 0.124715 & 0.129092 \\
        $10^{-5}$ & 0.127830 & 0.130582 & 0.128878 & 0.128369 \\
        $2\times10^{-5}$ & 0.136164 & 0.138593 & 0.135775 & 0.132319 \\
        \bottomrule
    \end{tabular}
\end{table}

\paragraph{Ordinal thresholds and boundary resolution.}
The ordinal thresholds also determine the coarse intervals in Eq.~\ref{eq:app-boundary-grid}. We compare the six-threshold configuration with a two-threshold alternative, $\{2/3,4/3\}$, and vary the boundary resolution $q\in\{2,3,4\}$. We select six thresholds with $q=2$ using validation MSE. Table~\ref{tab:scorer-boundary-resolution} shows that this configuration also gives the lowest mean test MSE. Increasing $q$ does not consistently improve regression; the highest mean Spearman correlation, obtained with six thresholds and $q=4$, exceeds the $q=2$ result by only 0.000079.

\begin{table}[h]
    \centering
    \caption{Ordinal-threshold and boundary-resolution comparison. Results are mean $\pm$ sample standard deviation over seeds 42 and 43. The six-threshold, $q=2$ row reuses the learning-rate baseline.}
    \label{tab:scorer-boundary-resolution}
    \begin{tabular}{lcrr}
        \toprule
        Ordinal thresholds & $q$ & MSE $\downarrow$ & Spearman $\rho_s$ $\uparrow$ \\
        \midrule
        Six thresholds & 2 & $\mathbf{0.123872}\pm0.002990$ & $0.811107\pm0.006701$ \\
        Six thresholds & 3 & $0.131291\pm0.000161$ & $0.805553\pm0.001183$ \\
        Six thresholds & 4 & $0.129176\pm0.003418$ & $\mathbf{0.811186}\pm0.002762$ \\
        Two thresholds & 2 & $0.134309\pm0.002946$ & $0.800777\pm0.009875$ \\
        Two thresholds & 3 & $0.131726\pm0.000921$ & $0.802823\pm0.008857$ \\
        Two thresholds & 4 & $0.133642\pm0.004875$ & $0.804636\pm0.008837$ \\
        \bottomrule
    \end{tabular}
\end{table}

\paragraph{Ordinal reweighting.}
Finally, we examine whether weighting ordinal thresholds or positive classes improves the scorer. In threshold order, we test four normalized weight configurations: uniform $(1,\dots,1)$, low-emphasis $(1.5, 1.3, \dots, 0.5)$, high-emphasis $(0.5, 0.7, \dots, 1.5)$, and middle-emphasis $(0.6, 1, 1.4, 1.4, 1, 0.6)$. These are crossed with four positive-class weight vectors:
\begin{equation*}
    \begin{aligned}
        \boldsymbol p_0&=(1,1,1,1,1,1),\\
        \boldsymbol p_1&=(0.823,1.037,1.212,1.291,1.450,2.016),\\
        \boldsymbol p_2&=(0.677,1.075,1.469,1.667,2.103,4.065),\\
        \boldsymbol p_3&=(0.458,1.155,2.159,2.780,4.422,16.524).
    \end{aligned}
\end{equation*}
Each combination uses five seeds. The final scorer retains unit weights based on validation MSE. Table~\ref{tab:scorer-ordinal-reweighting} reports the test results, where $\boldsymbol p_2$ with high-threshold emphasis reduces mean MSE from 0.127040 to 0.126756, a relative change of 0.22\%. The highest mean Spearman correlation instead occurs with $\boldsymbol p_3$ and uniform threshold weights (0.811850).

\begin{table}[t]
    \centering
    \caption{Ordinal reweighting: five-seed mean MSE. Columns indicate threshold-loss weighting; rows indicate positive-class weighting. The bold entry is the lowest mean.}
    \label{tab:scorer-ordinal-reweighting}
    \begin{tabular}{lrrrr}
        \toprule
        Positive-class weights & Uniform & Low & High & Middle \\
        \midrule
        $\boldsymbol p_0$ & 0.127040 & 0.131188 & 0.127591 & 0.132335 \\
        $\boldsymbol p_1$ & 0.128455 & 0.128698 & 0.127074 & 0.127770 \\
        $\boldsymbol p_2$ & 0.128300 & 0.131183 & \textbf{0.126756} & 0.127376 \\
        $\boldsymbol p_3$ & 0.128768 & 0.129774 & 0.128308 & 0.127996 \\
        \bottomrule
    \end{tabular}
\end{table}

\subsubsection{Final Scorer and Evaluation Protocol}
\label{app:final-scorer}

We retrain the fixed configuration in Table~\ref{tab:scorer-training-settings} with five random seeds. Within each run, we retain the checkpoint with the lowest validation MSE, then select the final checkpoint across runs using the same validation criterion. This selects seed 46 for all external ASR evaluations reported in Table~\ref{tab:system-level-comparison}. Table~\ref{tab:scorer-final-seeds} reports the test results for all five runs, separately from the development runs used in the preceding ablations. The resulting mean MSE is $0.127693\pm0.002318$, and mean Spearman correlation is $0.808080\pm0.004368$. The final scorer uses the fixed configuration rather than combining the best individual outcomes from different hyperparameter searches.

\begin{table}[h]
    \centering
    \caption{Independent retraining of the final scorer configuration. All results use the same 1,292-pair test split and raw predictions. Bold values identify the best result in each column.}
    \label{tab:scorer-final-seeds}
    \begin{tabular}{crrrr}
        \toprule
        Seed & MSE $\downarrow$ & MAE $\downarrow$ & Pearson $r$ $\uparrow$ & Spearman $\rho_s$ $\uparrow$ \\
        \midrule
        42 & 0.126302 & 0.254662 & 0.818037 & 0.807938 \\
        43 & 0.129235 & 0.255753 & 0.811441 & 0.804729 \\
        44 & 0.130549 & 0.250619 & 0.810363 & 0.804272 \\
        45 & 0.127689 & \textbf{0.244722} & 0.819069 & 0.808274 \\
        46 & \textbf{0.124689} & 0.245155 & \textbf{0.821666} & \textbf{0.815188} \\
        \bottomrule
    \end{tabular}
\end{table}

For external evaluation, we freeze the scorer and apply it to seven ASR systems on 647 FLEURS English utterances, 2,611 LibriSpeech test-clean utterances, and 1,830 VoxPopuli English utterances per system. The resulting 21 system--dataset combinations contain 35,616 transcript pairs in total. Each pair is scored independently in a single forward pass. We apply the clipping and normalization in Eq.~\ref{eq:normalized-scorer} to each prediction before taking the arithmetic mean within a system--dataset combination. These normalized means are the \name values reported in the main comparison table. The external datasets do not carry the constructed scalar targets used for scorer training, so the external analysis examines ASR-system behavior rather than reporting target-regression error.


\section{Detailed Comparison with AER}
\label{app:aer-comparison}

AER and \name both assess consequences of transcription errors beyond lexical mismatch. AER measures the fraction of reference-derived questions whose answers differ when conditioned on the reference and the ASR hypothesis~\citep{pulikodan25_interspeech}. \name instead assigns a situational distortion score directly to the transcript pair. We compare their system-level behavior using the 21 entries in Table~\ref{tab:system-level-comparison}. The AER entries come from the original paper's evaluation subsets, while the \name entries use the data described in Appendix~\ref{app:final-scorer}. The analysis focuses on model ordering, numerical separation between system means, and the computation required at scoring time.

\subsection{Agreement in Model Trends}

Both metrics assign a lower score to whisper-large than to whisper-tiny on all three datasets. They also favor wav2vec2-large over wav2vec2-base in every dataset, and both identify whisper-large as the best of the seven evaluated systems throughout. These shared preferences show that \name retains broad model-scaling trends also captured by the comprehension-based evaluation.

Agreement is strongest on FLEURS, where the two metrics produce exactly the same ordering of all seven systems. On LibriSpeech, 20 of the 21 pairwise orderings agree. The only reversal concerns whisper-small and whisper-medium. AER slightly favors whisper-small, with scores of 0.2600 and 0.2637, whereas \name favors whisper-medium, with scores of 0.1047 and 0.0818. The corresponding Spearman correlations between the seven system scores are 1.0000 on FLEURS and 0.9643 on LibriSpeech, as summarized in Table~\ref{tab:aer-score-comparison}.

The ordering differs more substantially on VoxPopuli. Across the five Whisper sizes, AER takes values 0.4723, 0.4505, 0.4567, 0.4739, and 0.4345, while \name decreases at every step from 0.3157 to 0.1684. Cross-family preferences also differ. For example, AER favors wav2vec2-large over whisper-base, with scores of 0.4366 and 0.4505, while \name favors whisper-base, with scores of 0.2532 and 0.3577. Consequently, the seven-system Spearman correlation on VoxPopuli is $-0.1071$. The shared preference for the largest model within each family therefore coexists with different finer-grained rankings on this dataset.

\begin{table}[htbp]
    \centering
    \caption{Score separation and rank agreement derived from the seven systems per dataset in Table~\ref{tab:system-level-comparison}. A span is the maximum score minus the minimum score. The span ratio is \name divided by AER, and $\rho_s$ is their Spearman rank correlation.}
    \label{tab:aer-score-comparison}
    \begin{tabular}{lrrrr}
        \toprule
        Dataset & AER span & \name span & Span ratio & $\rho_s$ \\
        \midrule
        FLEURS & 0.1001 & 0.2484 & 2.48 & 1.0000 \\
        LibriSpeech & 0.0469 & 0.1623 & 3.46 & 0.9643 \\
        VoxPopuli & 0.0394 & 0.2400 & 6.09 & $-0.1071$ \\
        \bottomrule
    \end{tabular}
\end{table}

\subsection{Separation Between System Scores}

Both metrics are reported on a $[0,1]$ scale, with lower values indicating less distortion. Within this range, \name uses a wider span to distinguish the evaluated systems on every dataset. On FLEURS, the seven AER values lie between 0.1477 and 0.2478, giving a span of 0.1001. The corresponding \name values range from 0.1388 to 0.3872, giving a span of 0.2484. The span is therefore 2.48 times larger while preserving exactly the same model ranking. This is the clearest case in Table~\ref{tab:system-level-comparison} of agreement in ordering accompanied by greater numerical separation.

The difference in span is larger on LibriSpeech and VoxPopuli. On LibriSpeech, AER occupies the interval from 0.2540 to 0.3009, whereas \name occupies 0.0634 to 0.2257, giving spans of 0.0469 and 0.1623. On VoxPopuli, all AER values are concentrated between 0.4345 and 0.4739, a span of 0.0394, while \name spans 0.1684 to 0.4084, a width of 0.2400. These produce the respective span ratios of 3.46 and 6.09 in Table~\ref{tab:aer-score-comparison}. Here, separation refers specifically to distances between reported system means on each metric's scale.

Several individual comparisons illustrate what this difference looks like in practice. On FLEURS, whisper-medium and whisper-large differ by only 0.0031 under AER, compared with 0.0202 under \name. On LibriSpeech, both metrics favor whisper-small over wav2vec2-large, but the difference is 0.0043 under AER and 0.0297 under \name. For wav2vec2-base and wav2vec2-large, the gaps are respectively 0.0092 versus 0.0423 on FLEURS, 0.0053 versus 0.0241 on LibriSpeech, and 0.0021 versus 0.0507 on VoxPopuli. Thus, several improvements that occupy a narrow numerical interval under AER are more visibly separated by \name.

The amount of separation nevertheless depends on the model pair. For whisper-medium and whisper-large on VoxPopuli, \name gives nearly equal scores of 0.1701 and 0.1684, whereas AER gives 0.4739 and 0.4345. Together with the ordering differences above, this shows that the wider overall spans coexist with selective compression of individual contrasts. \name's numerical separation reflects its own severity-based scoring rule rather than a uniform expansion of the AER values.

\subsection{Computation at Scoring Time}

AER's published workflow comprises question generation, separate answer generation from the reference and hypothesis, and comparison of the two answer sets~\citep{pulikodan25_interspeech}. Evaluating an uncached pair therefore involves four LLM calls. The reported setup uses Gemini 1.5 Flash for questions, GPT-4o-mini for both answer sets, and GPT-4o for judging. Questions and reference answers could be cached across hypotheses, but hypothesis answering and answer comparison would still require LLM inference.

For \name, LLM annotation and review are confined to offline supervision construction. Once trained, the DeBERTa-based scorer takes the reference and hypothesis together and returns a scalar in one forward pass. Scoring requires no question generation, answer generation, or LLM judgment. Evaluating another ASR system or checkpoint therefore adds only direct scorer passes over its transcript pairs. This removes repeated generative inference from routine evaluation while retaining the model-level trends and score separation described above.


\section{Controlled Dataset Construction and Human Annotation}
\label{app:human-annotation}

This appendix describes the controlled dataset $\mathcal{D}_{\mathrm{ctrl}}$ used in Section~\ref{sec:intrinsic-reliability}. We explain how fixed edits produce comparable hypothesis variants, illustrate the construction with a complete example from the dataset, and detail the human annotation and subsequent analyses.

\subsection{Dataset Composition and Controlled Construction}
\label{app:controlled-construction}

The dataset contains 30 English references, each paired with 11 hypothesis variants, yielding 330 distinct reference--hypothesis pairs. The references cover five domains, with the distribution shown in Table~\ref{tab:controlled-domain-counts}. Twenty-five references come from curated ASR examples, retaining two semantic edits from their ASR hypotheses. The remaining five references are designed scenarios with two specified semantic edits. Each reference also receives two synthetic expression-level changes, consisting of an immediate word repetition and a filled pause. Thus, the construction uses 120 fixed edits, comprising 50 ASR-derived semantic edits, 10 designed semantic edits, and 60 synthetic disfluencies.

\begin{table}[htbp]
    \centering
    \caption{Composition of the controlled dataset. Each reference has four candidate gaps for annotation and 11 hypothesis variants for evaluation. Culture includes cultural and sports content.}
    \label{tab:controlled-domain-counts}
    \begin{tabular}{lrrr}
        \toprule
        Domain & References & Candidate gaps & Transcript pairs \\
        \midrule
        Culture & 9 & 36 & 99 \\
        Finance & 4 & 16 & 44 \\
        Services & 7 & 28 & 77 \\
        Public affairs & 5 & 20 & 55 \\
        General communication & 5 & 20 & 55 \\
        \midrule
        Total & 30 & 120 & 330 \\
        \bottomrule
    \end{tabular}
\end{table}

At construction time, the four edits for each reference have provisional assignments of two $\mathcal{T}_0$ gaps, one $\mathcal{T}_1$ gap, and one $\mathcal{T}_2$ gap. These assignments specify the planned experimental combinations and are subsequently checked by human annotators. Starting from the unchanged reference, we apply selected subsets of the four edits to obtain the 11 profiles in Table~\ref{tab:controlled-example-profiles}. Every profile therefore occurs once for each reference. A profile requiring one $\mathcal{T}_0$ edit always uses the earlier of the two edits in reference character order; a profile requiring two uses both. The $\mathcal{T}_1$ and $\mathcal{T}_2$ edits likewise remain identical whenever they recur across profiles.

This construction holds the reference, surrounding context, and wording of each retained edit fixed. For example, the difference between a $\mathcal{T}_2$ variant and a $\mathcal{T}_2+\mathcal{T}_1$ variant is the addition of the same predefined $\mathcal{T}_1$ edit, while the $\mathcal{T}_2$ edit remains unchanged. Comparing variants within a reference therefore isolates specified changes in error composition. Comparing the same construction profile across references examines how the scorer responds to that composition in different contexts.

The edits are applied at recorded, nonoverlapping character spans. We verify that reversing all retained edits reconstructs the reference exactly, that the four-edit variant reproduces its parent hypothesis, and that all 330 reference--hypothesis pairs are unique. Gap identities are inherited from these controlled edits. The counts in this experiment refer to the recorded edit units, without rerunning the full CGD segmentation on every generated variant. The 30 references were retained from a pool of 35 using the prediction-based selection procedure described in Appendix~\ref{app:controlled-selection}.

\subsection{A Complete Construction Example}
\label{app:controlled-example}

Reference 27 concerns approval of a draft before sending it to a client. Its English reference is reproduced below.

\begin{quote}
I haven't approved the draft yet. The figures look right, but I still need to check the conclusion. Do not send it to the client until I approve it. Maya can keep working on the appendix in the meantime.
\end{quote}

Table~\ref{tab:controlled-example-edits} gives its four fixed edits. The two $\mathcal{T}_0$ edits introduce disfluencies, repeating ``I'' at the beginning and inserting ``uh,'' before ``meantime.'' Neither changes the approval status or the instruction about sending the draft. The $\mathcal{T}_1$ edit changes the assessment of the figures from ``The figures look right'' to ``Most of the figures look right.'' This changes what the listener understands about the figures, while preserving both the pending approval and the prohibition on sending the draft. The $\mathcal{T}_2$ edit reverses that prohibition, allowing the draft to be sent before approval. Both annotators agreed on all four assignments in this example.

\begin{table}[htbp]
    \centering
    \caption{The four fixed edits in reference 27. The symbol $\emptyset$ denotes an insertion. $G_0$ is inserted at the beginning of the reference, and $G_3$ immediately before ``meantime.'' Gap IDs indicate positions, independently of severity.}
    \label{tab:controlled-example-edits}
    \begin{tabular}{lp{0.31\linewidth}p{0.42\linewidth}c}
        \toprule
        Gap & Reference span & Hypothesis span & Tier \\
        \midrule
        $G_0$ & $\emptyset$ & I & $\mathcal{T}_0$ \\
        $G_1$ & The figures & Most of the figures & $\mathcal{T}_1$ \\
        $G_2$ & Do not send it to the client until I approve it. & You can send it to the client before I approve it. & $\mathcal{T}_2$ \\
        $G_3$ & $\emptyset$ & uh, & $\mathcal{T}_0$ \\
        \bottomrule
    \end{tabular}
\end{table}

The 11 variants are obtained by retaining the subsets in Table~\ref{tab:controlled-example-profiles}. The gap IDs in this table refer to the four parent edits, so that the same ID always identifies the same textual change. The suffixes $a$--$k$ identify the construction profiles used throughout the analysis. The example's predicted scores are included to show the actual output for each resulting pair.

\begin{table}[htbp]
    \centering
    \caption{All 11 variants of reference 27. Each hypothesis applies exactly the listed edits to the reference. The final column reports normalized \name predictions, rounded to four decimal places. These construction profiles are used for all 30 references.}
    \label{tab:controlled-example-profiles}
    \begin{tabular}{cllr}
        \toprule
        Variant & Construction profile & Retained edits & \name \\
        \midrule
        $a$ & $\mathcal{T}_0$ & $G_0$ & 0.0274 \\
        $b$ & $2\mathcal{T}_0$ & $G_0,G_3$ & 0.0564 \\
        $c$ & $\mathcal{T}_1$ & $G_1$ & 0.3204 \\
        $d$ & $\mathcal{T}_1+\mathcal{T}_0$ & $G_0,G_1$ & 0.3356 \\
        $e$ & $\mathcal{T}_1+2\mathcal{T}_0$ & $G_0,G_1,G_3$ & 0.3774 \\
        $f$ & $\mathcal{T}_2$ & $G_2$ & 0.7573 \\
        $g$ & $\mathcal{T}_2+\mathcal{T}_0$ & $G_0,G_2$ & 0.7751 \\
        $h$ & $\mathcal{T}_2+2\mathcal{T}_0$ & $G_0,G_2,G_3$ & 0.7952 \\
        $i$ & $\mathcal{T}_2+\mathcal{T}_1$ & $G_1,G_2$ & 0.7678 \\
        $j$ & $\mathcal{T}_2+\mathcal{T}_1+\mathcal{T}_0$ & $G_0,G_1,G_2$ & 0.7824 \\
        $k$ & $\mathcal{T}_2+\mathcal{T}_1+2\mathcal{T}_0$ & $G_0,G_1,G_2,G_3$ & 0.7934 \\
        \bottomrule
    \end{tabular}
\end{table}

For variant $c$, only $G_1$ is applied. Its complete hypothesis is
\begin{quote}
I haven't approved the draft yet. \textbf{Most of the figures} look right, but I still need to check the conclusion. Do not send it to the client until I approve it. Maya can keep working on the appendix in the meantime.
\end{quote}

For variant $f$, only $G_2$ is applied. The figure assessment returns to the reference wording, and the permission to send the draft changes instead.
\begin{quote}
I haven't approved the draft yet. The figures look right, but I still need to check the conclusion. \textbf{You can send it to the client before I approve it.} Maya can keep working on the appendix in the meantime.
\end{quote}

For variant $k$, all four edits are retained.
\begin{quote}
\textbf{I} I haven't approved the draft yet. \textbf{Most of the figures} look right, but I still need to check the conclusion. \textbf{You can send it to the client before I approve it.} Maya can keep working on the appendix in the \textbf{uh,} meantime.
\end{quote}

The reference and each selected change remain verbatim across these variants. Thus, moving from $f$ to $i$ adds the same change in the figure assessment shown in $c$, and moving from $i$ to $k$ adds the two specified disfluencies. The variants form a controlled set because their differences are known and reusable at fixed positions within the same situation.

\subsection{Annotation Rules and Procedure}
\label{app:human-procedure}

Two PhD students independently annotated the 120 candidate gaps. Each received 30 records containing the complete English reference, the hypothesis with all four changes, and a list of the four candidate gaps. Each candidate identifies its changed span within the full reference context and has an initially empty \texttt{level} field. The annotators filled this field with \texttt{T0}, \texttt{T1}, or \texttt{T2}, preserving the record IDs, gap IDs, and transcript text. The task assigns nominal tiers only; it does not ask annotators to supply numerical distortion scores or directional boundary adjustments.

The instructions require the annotator to first read the complete reference and establish what it describes, communicates, or seeks to accomplish. Each gap is then judged through its single-gap hypothesis, with the other three edits restored to the reference. In the preceding example, judging $G_1$ means comparing the reference with variant $c$. The permission reversal in $G_2$ is absent from that comparison, so its consequences do not determine the label of $G_1$.

The three annotation rules are as follows.
\begin{enumerate}
    \item $\mathcal{T}_0$ applies when expression changes while the conveyed situation, understanding, and communicative purpose remain intact. A filler or immediate repetition is assigned to this tier when it introduces a disfluency without changing the relevant information or instructions.
    \item $\mathcal{T}_1$ applies when the listener's understanding of some information changes but the intended outcome remains preserved. The annotator identifies both the information that changes and the purpose or requirement that survives. In reference 27, the assessment of the figures changes while approval remains pending and sending remains prohibited.
    \item $\mathcal{T}_2$ applies when the change affects the intended outcome, including what judgment should be made, what action should be executed, or what result is expected. In reference 27, the altered instruction permits an action that the reference explicitly prohibits before approval.
\end{enumerate}

The instructions explicitly discourage assigning severity from the number of changed words, the mere presence of a lexical error, or the fluency of the hypothesis. A short edit may change an execution requirement, whereas a longer expression change may preserve the situation. The same lexical replacement may also receive different labels in different reference contexts. Annotators are not required to reproduce the planned allocation of two $\mathcal{T}_0$, one $\mathcal{T}_1$, and one $\mathcal{T}_2$ gap. If a gap cannot be judged, the instructions allow its label to remain empty with the record and gap IDs flagged for review. Both returned annotations contained valid labels for all 120 gaps.

The annotation unit is the individual parent gap. Once its tier is assigned, that label is carried into every hypothesis variant containing the same edit. Annotators therefore make 120 judgments each, and these judgments supply labels for the 330 derived pairs. Before computing agreement, we align the two returned files by reference and gap IDs and verify that their reference text, hypothesis text, and marked contexts match. When mapping labels to derived variants, we match the reference and the full marked edit context, allowing the local gap numbering to change.

\subsection{Annotation Agreement and Final Labels}
\label{app:human-agreement-details}

\textbf{Agreement before adjudication.}
The two annotators agree on 108 of the 120 gaps, giving 90.00\% exact agreement. Table~\ref{tab:human-confusion} shows their full label counts. The 12 disagreements occur across nine references. Seven concern the $\mathcal{T}_1$--$\mathcal{T}_2$ boundary, three concern $\mathcal{T}_0$--$\mathcal{T}_1$, and two concern $\mathcal{T}_0$--$\mathcal{T}_2$. Thus, 10 of the 12 disagreements involve adjacent tiers. The largest group concerns whether an information change preserves or alters the intended outcome.

\begin{table}[htbp]
    \centering
    \caption{Human annotation confusion matrix before adjudication. Rows give annotator A's labels and columns give annotator B's labels for the same 120 gaps.}
    \label{tab:human-confusion}
    \begin{tabular}{lrrrr}
        \toprule
        & B $\mathcal{T}_0$ & B $\mathcal{T}_1$ & B $\mathcal{T}_2$ & Total \\
        \midrule
        A $\mathcal{T}_0$ & 58 & 0 & 0 & 58 \\
        A $\mathcal{T}_1$ & 3 & 24 & 3 & 30 \\
        A $\mathcal{T}_2$ & 2 & 4 & 26 & 32 \\
        \midrule
        Total & 63 & 28 & 29 & 120 \\
        \bottomrule
    \end{tabular}
\end{table}

Cohen's $\kappa$ is 0.8396. Treating the tiers as ordinal categories gives linearly weighted $\kappa=0.8675$ and quadratically weighted $\kappa=0.8924$, using disagreement weights proportional to the absolute and squared tier distances, respectively. Their increase over unweighted $\kappa$ reflects the predominance of adjacent-tier disagreements. These statistics use the two independent human label sets before any resolution with model labels.

At the transcript-pair level, exact agreement requires the two annotators to assign the same label to every retained gap in a variant. This holds for 268 of 330 pairs, or 81.21\%, as reported in Table~\ref{tab:human-agreement}. One disputed parent gap appears in several variants, and a variant with multiple gaps must satisfy all corresponding label matches. Consequently, the lower pair-level agreement is a stricter summary of the same 120 judgments. If agreement is instead defined only by the counts of each tier, 274 pairs agree, or 83.03\%. The six additional matches exchange the $\mathcal{T}_1$ and $\mathcal{T}_2$ assignments of two gaps while retaining the same tier counts. The main text uses the stricter gap-identity criterion.

\textbf{Resolution of disagreements.}
Whenever the two human labels agree, their label is adopted. Otherwise, the initial model-assigned tier provides a third vote. It agrees with annotator A on five of the 12 disputed gaps and with annotator B on seven, resolving all disagreements without a three-way tie. This third vote is a categorical gap label from construction, distinct from the continuous \name prediction evaluated in Section~\ref{sec:intrinsic-reliability}. The final labels contain 60 $\mathcal{T}_0$, 31 $\mathcal{T}_1$, and 29 $\mathcal{T}_2$ gaps.

The final labels differ from the provisional construction assignments at one gap in reference 13. Both human annotators assign $\mathcal{T}_1$ to a gap initially assigned $\mathcal{T}_2$, so the human agreement overrides that initial label. All six variants containing the gap inherit this correction. The underlying text and predicted scores stay the same. Consequently, the 11 construction-profile groups in Table~\ref{tab:intrinsic-score-structure} each retain 30 pairs, while their profile names record the original construction assignments. The within-reference ranking analysis uses the final adjudicated labels of every retained gap.

\subsection{Ranking Protocol and Detailed Results}
\label{app:human-ranking}

For a hypothesis $h$ belonging to reference $r$, let $c_k(r,h)$ be the number of retained gaps whose final label is $\mathcal{T}_k$. Its expected severity order is defined by the count vector
\[
    \mathbf{v}(r,h)=\bigl(c_2(r,h),c_1(r,h),c_0(r,h)\bigr).
\]
We compare these vectors lexicographically, prioritizing $\mathcal{T}_2$ counts, followed by $\mathcal{T}_1$ counts and then $\mathcal{T}_0$ counts. Equal vectors receive equal expected ranks. This evaluates the tier-priority and count-refinement behavior of the scorer using the final nominal labels. The expected ordering is derived from those labels under the stated aggregation rule; annotators do not separately rank all 330 hypotheses.

The predicted score is normalized as
\[
    \widehat{s}(r,h)
    =\frac{\min\{7/3,\max\{0,f_\theta(r,h)\}\}}{7/3}.
\]
All 330 raw predictions fall within the clipping bounds. For each reference, we compare all $\binom{11}{2}=55$ pairs of variants. If their count vectors differ, agreement requires the more severe variant to have a strictly higher predicted score. A predicted tie does not satisfy this strict ordering. Pairs with equal expected count vectors are excluded from pairwise agreement, while Spearman's $\rho_s$ includes all 11 variants using average ranks for ties.

The correction in reference 13 changes its expected order to
\[
    a<b<c=f<d=g<e=h<i<j<k.
\]
Its three expected ties, $(c,f)$, $(d,g)$, and $(e,h)$, reduce the number of strict comparisons for that reference from 55 to 52. The other 29 references each contribute 55, giving $29\times55+52=1{,}647$ comparisons. If $K_r$ of the $M_r$ eligible comparisons agree for reference $r$, the main-text agreement is
\[
    \mathrm{Agr.}=\frac{1}{30}\sum_{r=1}^{30}\frac{K_r}{M_r}.
\]
Spearman correlations are also computed separately for each reference and then averaged. Domain results use the same averaging rule over references in that domain.

\textbf{Overall and domain results.}
The mean per-reference agreement is 84.07\%, with a median of 85.45\% and a mean Spearman correlation of 0.8001. In total, 1,385 of the 1,647 strict comparisons agree. Pooling all comparisons directly gives 84.09\%; its small difference from the reported 84.07\% follows from the unequal denominators of 52 and 55. Across domains, mean agreement ranges from 80.00\% for services to 88.48\% for culture, with mean $\rho_s$ ranging from 0.7078 to 0.8758, as detailed in Table~\ref{tab:intrinsic-ranking}. These results summarize consistency across 30 reference groups, whose variants share text and edits.

\textbf{Where ordering succeeds and fails.}
To locate the remaining ordering errors, Table~\ref{tab:controlled-ranking-by-tier} groups each strict comparison by the highest tier at which its two count vectors differ. A $\mathcal{T}_1$ comparison therefore has identical $\mathcal{T}_2$ counts but different $\mathcal{T}_1$ counts. A $\mathcal{T}_0$ comparison has identical counts at both higher tiers and differs only in disfluency count. Each comparison belongs to exactly one group. The percentages in this table pool comparisons within each group.

\begin{table}[htbp]
    \centering
    \caption{Ordering agreement by the highest tier with a count difference. Counts use final adjudicated labels. Agreement is the fraction of correct strict comparisons within each row.}
    \label{tab:controlled-ranking-by-tier}
    \begin{tabular}{lrrr}
        \toprule
        Distinguishing tier & Comparisons & Correct & Agreement \\
        \midrule
        $\mathcal{T}_2$ & 870 & 793 & 91.15\% \\
        $\mathcal{T}_1$ & 471 & 393 & 83.44\% \\
        $\mathcal{T}_0$ & 306 & 199 & 65.03\% \\
        \midrule
        Total & 1,647 & 1,385 & 84.09\% \\
        \bottomrule
    \end{tabular}
\end{table}

Agreement is highest when the variants differ in $\mathcal{T}_2$ count and lowest when only $\mathcal{T}_0$ count differs. Of the 262 incorrect strict comparisons, 107 concern only the number of disfluencies. The scorer thus reproduces the larger severity distinctions more consistently than the smaller refinements from expression-level accumulation. Reference 27 illustrates this distinction. Its pure $\mathcal{T}_0$, $\mathcal{T}_1$, and $\mathcal{T}_2$ variants receive scores of 0.0274, 0.3204, and 0.7573. However, variant $h$ scores 0.7952, above $i$, $j$, and $k$, despite the latter retaining the additional $\mathcal{T}_1$ edit and having higher expected ranks. This example shows both the separation of basic tiers and a failure to preserve their full compositional ordering.

\subsection{Reference Selection and Scope of the Analysis}
\label{app:controlled-selection}

The final 30 references were retained from 35 candidate references after inspecting the existing \name predictions. The recorded selection criterion maximizes the mean-score separation between the $\mathcal{T}_2$ construction family ($f,g,h$) and the $\mathcal{T}_1$ construction family ($c,d,e$), standardized by the square root of the mean of the six within-profile sample variances. Five references were excluded through this prediction-based selection. Their exclusion was not established independently as an annotation-quality correction.

Accordingly, the score distributions and ranking results describe this selected diagnostic dataset. In particular, the separation of the $\mathcal{T}_1$ and $\mathcal{T}_2$ profile families is influenced by the selection criterion and is not an independent estimate for randomly sampled references. Human agreement is measured separately on the retained gaps before adjudication, while final ranking targets incorporate the initial model label only when the two humans disagree. The controlled edits make individual score changes inspectable, and the reported counts specify the evidence available at the reference, gap, and derived-pair levels.

\section{Case Study}
\label{app:case-study}
We examine two types of transcription variation: deletion of a negation at different informational positions, and accumulation of expression-level disfluencies. Tables below report BERTScore$_{\mathrm{F1}}$ (higher is better), WER (lower is better), and \name (Ours; lower indicates less distortion). Complete references and hypotheses accompany the scores. Highlighted words identify the relevant negations or altered expressions; deleted words are omitted from the hypotheses.

\subsection{Negation Deletion in Different Contexts}
\label{app:case-negation}
Each hypothesis deletes one occurrence of \emph{not}. N0 changes the stated reason for a flight choice while preserving the booking request; N1 changes the explicit flight choice and creates a conflict with the request to book the earlier flight. N2 reverses lunch availability while leaving meeting attendance intact; N3 reverses meeting attendance itself. These examples use separate references, reproduced in full below.

\begin{center}
\small
\renewcommand{\arraystretch}{1.18}
\begin{tabular}{@{}lp{0.40\linewidth}rrr@{}}
\toprule
Index & Change & BERTScore$_{\mathrm{F1}}$ & WER & Ours \\
\midrule
N0 (flight) & Reason reversed & 0.9866 & 0.0526 & 0.6815 \\
N1 (flight) & Core flight choice reversed & 0.9668 & 0.0526 & \textbf{0.7114} \\
N2 (meeting) & Secondary availability reversed & 0.9514 & 0.0526 & 0.8006 \\
N3 (meeting) & Meeting attendance reversed & 0.9534 & 0.0526 & \textbf{0.8293} \\
\bottomrule
\end{tabular}
\end{center}
WER assigns the same value to all four examples. Ours assigns a higher score to the core-event change within each scenario: N0 $<$ N1 and N2 $<$ N3. BERTScore agrees with this ordering for the flight pair, but reverses it for the meeting pair. Thus, the meeting examples illustrate how a similar surface edit can have different situational consequences that lexical error counts and representation similarity do not necessarily distinguish. The Ours values in this table are normalized using Eq.~\ref{eq:normalized-scorer}.

\begingroup
\definecolor{casechange}{HTML}{B3261E}
\newcommand{\casechange}[1]{\textcolor{casechange}{\textbf{#1}}}
\begin{sitarcase}
\noindent\textbf{N0: Reason reversed.}\par
\noindent\textbf{Ref:} I chose the 8:15 flight because I am \casechange{not} worried about the price. Please book the 8:15 refundable ticket.\par
\smallskip\noindent\textbf{Hyp:} I chose the 8:15 flight because I am worried about the price. Please book the 8:15 refundable ticket.\par
\end{sitarcase}
\begin{sitarcase}
\noindent\textbf{N1: Core flight choice reversed.}\par
\noindent\textbf{Ref:} The 8:15 flight is available. I have decided \casechange{not} to take the 9:40 flight. Please book the earlier one.\par
\smallskip\noindent\textbf{Hyp:} The 8:15 flight is available. I have decided to take the 9:40 flight. Please book the earlier one.\par
\end{sitarcase}
\begin{sitarcase}
\noindent\textbf{N2: Secondary availability reversed.}\par
\noindent\textbf{Ref:} The meeting is on Thursday at three. Maya is \casechange{not} available for lunch, but she will attend the meeting.\par
\smallskip\noindent\textbf{Hyp:} The meeting is on Thursday at three. Maya is available for lunch, but she will attend the meeting.\par
\end{sitarcase}
\begin{sitarcase}
\noindent\textbf{N3: Meeting attendance reversed.}\par
\noindent\textbf{Ref:} The meeting is on Thursday at three. Maya is available for lunch, but she will \casechange{not} attend the meeting.\par
\smallskip\noindent\textbf{Hyp:} The meeting is on Thursday at three. Maya is available for lunch, but she will attend the meeting.\par
\end{sitarcase}

\endgroup

\subsection{Accumulating Expression-Level Redundancy}
\label{app:case-disfluency}
F0--F4 share one reference and progressively introduce inflectional variation, fillers, repetitions, a lengthened spelling, and punctuation changes. The show time and the question about reserving seats remain unchanged. The table summarizes the cumulative edits; the complete texts retain the original punctuation and wording.

\begin{center}
\small
\renewcommand{\arraystretch}{1.18}
\begin{tabular}{@{}lp{0.47\linewidth}rrr@{}}
\toprule
Index & Cumulative change & BERTScore$_{\mathrm{F1}}$ & WER & Ours \\
\midrule
F0 & Inflection: starts $\rightarrow$ start & 0.9783 & 0.0345 & 0.112 \\
F1 & F0 + just repetition + comma deletion & 0.9404 & 0.1034 & 0.281 \\
F2 & F1 + filler Um & 0.9428 & 0.1724 & 0.317 \\
F3 & F2 + I repetition + reaaally + comma deletion & 0.8976 & 0.2414 & \textbf{0.330} \\
F4 & F3 + filler uh + the repetition + further punctuation removal & 0.8611 & 0.3793 & \textbf{0.321} \\
\bottomrule
\end{tabular}
\end{center}
WER rises from 0.0345 to 0.3793, while BERTScore decreases overall from 0.9783 to 0.8611. Ours initially rises from 0.112 to 0.330 and then changes slightly to 0.321 despite the additional surface edits in F4. This pattern illustrates a limited response to accumulating expression-level variation, with a small final decrease rather than strict monotonicity.

\begingroup
\definecolor{casechange}{HTML}{B3261E}
\newcommand{\casechange}[1]{\textcolor{casechange}{\textbf{#1}}}
\begin{sitarcase}
\noindent\textbf{Shared reference}\par
Well, I think the show starts at eight, and it sounds really good. I just want to know whether we need to reserve seats before we get on board.\par
\end{sitarcase}
\begin{sitarcase}
\noindent\textbf{Hyp-F0}\par
Well, I think the show \casechange{start} at eight, and it sounds really good. I just want to know whether we need to reserve seats before we get on board.\par
\end{sitarcase}
\begin{sitarcase}
\noindent\textbf{Hyp-F1}\par
Well, I think the show \casechange{start} at eight and it sounds really good. I \casechange{just just} want to know whether we need to reserve seats before we get on board.\par
\end{sitarcase}
\begin{sitarcase}
\noindent\textbf{Hyp-F2}\par
\casechange{Um, well,} I think the show \casechange{start} at eight and it sounds really good. I \casechange{just just} want to know whether we need to reserve seats before we get on board.\par
\end{sitarcase}
\begin{sitarcase}
\noindent\textbf{Hyp-F3}\par
\casechange{Um, well I I} think the show \casechange{start} at eight and it sounds \casechange{reaaally} good. I \casechange{just just} want to know whether we need to reserve seats before we get on board.\par
\end{sitarcase}
\begin{sitarcase}
\noindent\textbf{Hyp-F4}\par
\casechange{Um uh well I I} think \casechange{the the} show \casechange{start} at eight and it sounds \casechange{reaaally} good I \casechange{just just} want to know whether we need to reserve seats before we get on board\par
\end{sitarcase}

\endgroup

\end{document}